\documentclass[runningheads]{llncs}

\usepackage{macro}
\begin{document}

\title{The \sys Bound Propagator for Formal Analysis of Neural Networks}
\titlerunning{The \sys Bound Propagator for Formal Analysis of Neural Networks}

\author{Henry LeCates\inst{}\orcidlink{0009-0008-2374-4295}\thanks{Corresponding author.} \and
        Haoze Wu\orcidlink{0000-0002-5077-144X}}
\institute{Amherst College, USA\\
\email{\{hlecates26,hwu\}@amherst.edu}}

\maketitle        

\begin{abstract}

The parameterized CROWN analysis, a.k.a.,
\alphaCROWN, has emerged as a practically successful abstract interpretation method for neural network verification. However, existing implementations of \alphaCROWN are limited to Python, which complicates integration into existing DNN verifiers and long-term production-level systems. We introduce \sys, a new abstract-interpretation-based bound propagator implemented in C++. \sys supports Interval Bound Propagation, the DeepPoly/CROWN analysis, and the \alphaCROWN analysis over a general computational graph. We describe the architecture of \sys and show that it outperforms the state-of-the-art \alphaCROWN implementation in terms of both bound tightness and computational efficiency on supported benchmarks from VNN-COMP 2025. \sys is publicly available at \url{https://github.com/ai-ar-research/luna}.

\end{abstract}

\begin{comment}

The parameterized CROWN analysis, a.k.a.,
alpha-CROWN, has emerged as a practically successful abstract interpretation method for neural network verification. However, existing implementations of alpha-CROWN are limited to Python, which complicates integration into existing DNN verifiers and long-term production-level systems. We introduce Luna, a new abstract-interpretation-based bound propagator implemented in C++. Luna supports
Interval Bound Propagation, the DeepPoly/CROWN analysis, and the alpha-CROWN analysis over a general computational graph. We describe the architecture of Luna and show that it outperforms the state-of-the-art alpha-CROWN implementation in terms of both bound tightness and computational efficiency on supported benchmarks from VNN-COMP 2025. Luna is publicly available at https://github.com/ai-ar-research/luna.
\end{comment}
\section{Introduction}

Bound propagation based on abstract interpretation is an important technique in the formal analysis of deep neural networks (DNNs). Modern DNN reasoners, such as $\alpha$-$\beta$-CROWN~\cite{zhang2022general,wang2021beta,xu2020automatic}, CoRA~\cite{ladner2024exponent},
ERAN~\cite{kpoly,deeppoly,singh2019boosting}, 
Marabou~\cite{katz2019marabou,wu2024marabou,wu2022efficient}, 
NeuralSAT~\cite{duong2025neuralsat},
NNV~\cite{tran2020nnv,nnvTwo}, nnenum~\cite{nnenum}, PyRAT~\cite{lemesle2024neural}, VeriNet~\cite{verinet,deepsplit}, etc, all employ some version of abstract interpretation, either using it as a stand-alone, incomplete verification procedure, or as an in-processing analysis within a branch-and-bound or case-analysis search shell.

Over the past few years, a variety of bound propagation methods for DNNs have been proposed (e.g.,~\cite{wang2018efficient,deeppoly,zhang2018efficient,prima,zelazny2022reducing,huang2026parameterized,wu2022scalable,wei2023convex}, inter alia). These methods differ in the choice of the abstraction domain and abstraction refinement strategies, and span over the spectrum of the efficiency and precision trade-off, from exact bounding based on MILP~\cite{mipverify}, to fast bounding based on interval arithmetic~\cite{gowal2018effectiveness}. 

In this paper, we focus on the parameterized CROWN analysis (a.k.a. $\alpha$-CROWN)~\cite{zhang2018efficient}, which empirically strikes a good balance between precision and efficiency, and underlies the top two performers ($\alpha$-$\beta$-CROWN and NeuralSAT) in the past two VNN-COMPs~\cite{brix2024fifth,kaulen2025}. To date, practical implementations of the $\alpha$-CROWN abstract transformers are only available in Python. The most widely used implementation is \autolirpa~\cite{xu2020automatic}, which serves as the analysis backend for both $\alpha$-$\beta$-CROWN and NeuralSAT. While \autolirpa is itself a high-performance tool, invoking the Python-based $\alpha$-CROWN from systems written in other programming languages, e.g., C++ or MATLAB, can incur nontrivial integration and start-up overhead.

To reduce this barrier and make $\alpha$-CROWN accessible as a more easily integrable component for a wide range of verification tools, we develop \sys, the first C++ implementation of $\alpha$-CROWN with a stable interface suitable for foreign-function integration. The main contribution of this paper is to present this new abstract-interpretation-based tool, which is designed for the common setting in which a verifier must obtain sound output bounds of a neural network for a given input region. %\sys incurring minimal per-invocation overhead and still leveraging parallelization when available.
\sys currently includes the following features:
\begin{itemize}
    \item support for interval bound propagation (IBP), DeepPoly/CROWN, and \alphaCROWN
analyses;
    \item parallelization for the core tensor and bound computations leveraging the Torch Deep Learning Library;
    \item multiple invocation modes, including a command-line interface, a native C++ API, and Python bindings.
\end{itemize}

Beyond presenting the tool itself, this paper documents how \sys navigates the many implementation-level design choices that the \alphaCROWN algorithm leaves unspecified. These choices are rarely discussed in the literature, yet they meaningfully affect both precision and efficiency. We describe \sys's concrete decisions and their rationale in Section~\ref{sec:analysis-engine}.

On benchmarks from VNN-COMP~2025, \sys produces comparable or tighter output bounds than \autolirpa, the state-of-the-art $\alpha$-CROWN implementation, with overall lower runtime and start-up overhead. We expect that \sys will lower the engineering effort required to integrate strong bound propagation techniques into existing tools and to promote research work that focus on other algorithmic bottlenecks in neural network verification.
\section{Preliminaries}
Let $f:\mathbb{R}^n \to \mathbb{R}^m$ be a feedforward neural network. Given vectors $\ell,u \in \mathbb{R}^n$, define the input box $\mathcal{X} := \{ x \in \mathbb{R}^n \mid \ell \le x \le u \}$. The goal of bound propagation is to compute an output box $\mathcal{Y} := \{ y \in \mathbb{R}^m \mid a \le y \le b \}$ such that $f(x) \in \mathcal{Y}$ for all $x \in \mathcal{X}$. % We briefly review the two abstract domains relevant to this paper. In the following sections we follow the \alphaCROWN literature in using $\alpha$ to denote relaxation parameters. 

\subsubsection*{CROWN}

The CROWN analysis~\cite{zhang2018efficient}, or equivalently, the DeepPoly analysis~\cite{deeppoly}, operates by constructing affine relaxations of each nonlinear layer.
For a nonlinear layer $z' = \sigma(z)$ where $z \in [\ell,u]$,
it builds affine functions $h_L(z)$ and $h_U(z)$ such that
$h_L(z) \le \sigma(z) \le h_U(z) \text{ for all } z \in [\ell,u]$.
By composing these affine bounds layer by layer, one obtains
global affine lower and upper bounds on the network output.
The concrete output bounds are then obtained by
concretizing the global affine bounds using the input box $\mathcal{X}$.

\begin{figure}[t]
  \centering
  \begin{tikzpicture}
  \begin{axis}[
      width=0.48\textwidth,
      height=0.38\textwidth,
      xtick=\empty,
      ytick=\empty,
      tick style={draw=none},
      axis lines=middle,
      axis line style={->},
      xmin=-4.5, ymin=-4, xmax=5.5, ymax=5,
      samples=100,
      clip=false,
      ylabel={\({\scriptsize \text{ReLU}(z)}\)},
      xlabel={\({\scriptsize z}\)},
      every axis x label/.style={at={(ticklabel* cs:1)}, anchor=west},
      every axis y label/.style={
          at={(current axis.north)},
          anchor=south
      },
  ]
\addplot[gray, thin, dashed] coordinates {(-3,-3.4) (-3,0)};
\addplot[gray, thin, dashed] coordinates {(4,-3.4) (4,4)};
  \addplot[black, thick, samples at={-3, 0, 4}] {max(0,x)};
  \addplot[blue, thick, dashed, domain=-3:4] {(4/7)*(x + 3)};
  \addplot[red, thick, dashed, domain=-3:4] {0};
  \addplot[red, thick, dashed, domain=-3:4] {0.33*x};
  \addplot[red, thick, dashed, domain=-3:4] {0.66*x};
  \addplot[red, thick, dashed, domain=-3:4] {x};
  \node[below, font=\small] at (axis cs:-3,-3.4) {$l$};
  \node[below, font=\small] at (axis cs: 4,-3.4) {$u$};
  \end{axis}
  \end{tikzpicture}
  \caption{Family of valid linear relaxations for an unstable ReLU neuron with pre-activation bounds $[l,u]$. The lower bound can be given by any line across the origin with slope $\alpha \in [0,1]$. Standard \CROWN fixes $\alpha$ heuristically, while \alphaCROWN treats $\alpha$ as an optimizable parameter.}
  \label{fig:relu_relaxation_prelim}
\end{figure}

\subsubsection*{Parameterized CROWN ($\alpha$-CROWN)}

Many nonlinearities admit multiple valid affine relaxations.
Consider the ReLU activation function. As illustrated in Figure~\ref{fig:relu_relaxation_prelim}, $\mathrm{ReLU}(z) = \max(z, 0) \ge \alpha \cdot z \text{ for all } \alpha \in [0,1]$.
Instead of fixing the relaxation, $\alpha$-CROWN~\cite{xu2020automatic} considers a family of
linear relaxations parameterized by a variable $\alpha$.
This induces parameterized global bounds
$L(x,\alpha) \le f(x) \le U(x,\alpha)$ where $L(x,\alpha)$ and $U(x,\alpha)$ are affine in $x$ for fixed $\alpha$. 
The choice of $\alpha$ is then optimized (once for lower bound and once for upper bound) by solving
\[
\max_{\alpha \in \mathcal{A}} 
\;
\min_{x \in \mathcal{X}} L(x,\alpha),
\qquad
\min_{\alpha \in \mathcal{A}} 
\;
\max_{x \in \mathcal{X}} U(x,\alpha),
\]
where $\mathcal{A}$ represents the valid ranges of $\alpha$ (e.g., $[0,1]$ for ReLU lower bound). In practice, this nonconvex optimization problem is solved using projected gradient descent~\cite{madry2017towards} over $\alpha$. Note that for ReLU, $\alpha$ parameterizes only the lower-bound relaxation slope; the relaxation's upper bound is fixed. The concrete output upper bound $U(x,\alpha)$ nonetheless depends on $\alpha$ because, during back-substitution, the sign of the accumulated coefficient at each neuron selects either the neuron's lower or upper relaxation. A lower-bound slope therefore enters the output upper-bound expression whenever the corresponding coefficient is negative, which is why the upper-bound pass still optimizes $\alpha$. %Further details on the parameterized transformers and the optimization procedure are in Appendix~\ref{app:alpha-crown}.
\section{The \sys Bound Propagator}
\label{sec:methodology}

\begin{figure}[t]
  \centering
  \includegraphics[width=1.0\textwidth]{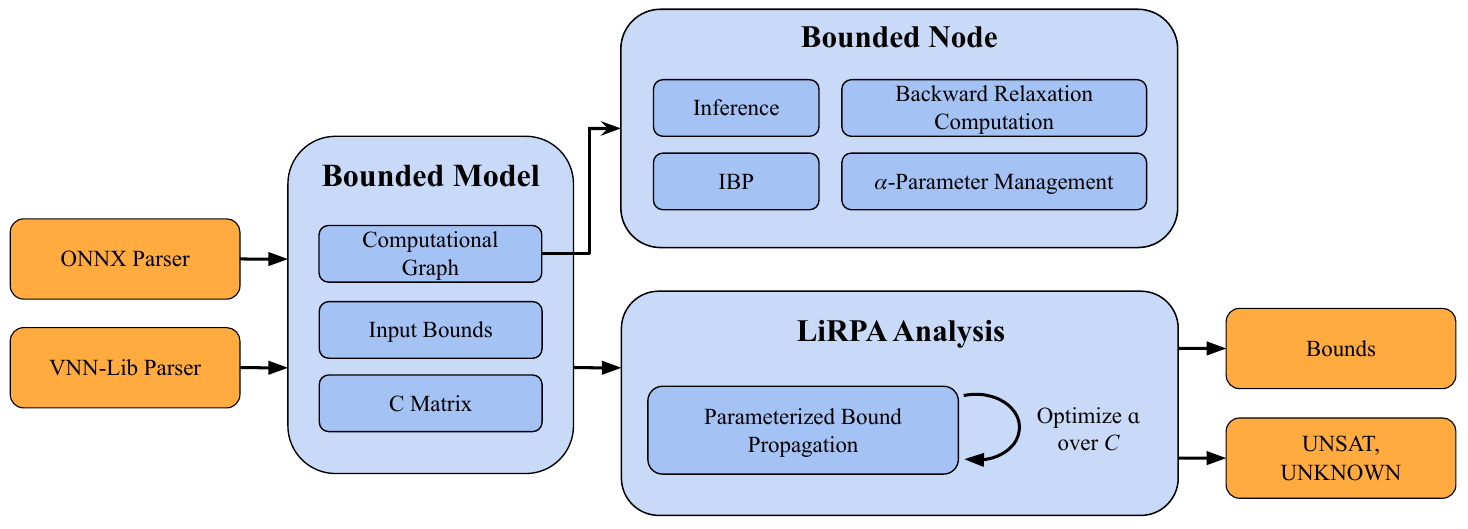}
  \caption{High-level overview of \sys's system architecture.}
  \label{fig:architecture}
\end{figure}

This section describes the design and architecture of \sys. A high-level overview is presented in Figure~\ref{fig:architecture}. \sys takes as input a neural network in the ONNX format~\citep{bai2019} and a specification in the VNN-LIB format, and computes sound bounds with respect to the input region defined in the specification. We now describe components of \sys.

\subsection{Bounded Model}

\sys parses the inputs and converts them into (i) a bound-aware computational graph on top of the Torch Deep Learning Library~\citep{paszke2019pytorch}, (ii) an input domain with element-wise lower and upper bounds, and (iii) a set of linear output constraints in the form $C y\le t$. These three components constitute the Bounded Model, which is an internal representation of the verification problem shared by all analysis routines. 

The graph-structure is stored as a directed acyclic graph (DAG), whose nodes correspond to layer operations. The DAG representation is necessary as neural networks can contain non-sequential architectures, such as residual (i.e., skip) connections and multi-branch structures. \sys's DAG maintains two maps containing the forward and backward dependencies of each node. These dual maps allow for efficient forward traversal during Interval Bound Propagation (IBP), and backward substitution during \CROWN. Each layer can maintain three bounds: (i) concrete numerical bounds from IBP, (ii) linear relaxations from \CROWN, and (iii) optimized bounds from \alphaCROWN.

The specification matrix $C$ represents a set of linear output constraints on the network's outputs. During bound propagation, \sys appends $C$ as an additional affine layer $h(y) = Cy$ at the end of the network, so the backward pass computes bounds on $Cy$ rather than on the raw network outputs $y$. Since this layer is affine, it is handled exactly during back substitution with no relaxation required. For example, for the output constraint $y_1 \leq y_2$, \sys seeks to bound the value of an auxiliary variable $a = y_1 - y_2$. If no output specification is provided in the VNN-LIB file, an identity matrix is constructed and treated as the output layer, allowing each output dimension to be bounded independently. Disjunctive (OR) constraints are handled by encoding each disjunct as a separate branch, composed into a single matrix that serves as the output layer. \sys then maintains maps of disjuncts and the corresponding indices; verification succeeds if any branch is satisfied. 

Like existing state-of-the-art neural network verification tools, \sys represents all bounds as floating-point tensors and computes with them in floating-point arithmetic, so its soundness guarantees hold modulo floating-point rounding.

\subsection{Bounded Node Modules}

Each operation in the ONNX network is represented as a Bounded Node object, which implements operation-specific bound computation. Each node implements two types of bounding: (i) Interval Bound Propagation (IBP), used before \CROWN and \alphaCROWN analyses to obtain initial bounds and detect stable neurons; and (ii) CROWN backward substitution, which takes as input the current affine bound coefficients, substitutes them backward, and outputs the linear bounds of the output in terms of the current node. Current supported operations are shown in Table~\ref{tab:supported-ops}. 

Adding a new operation to \sys requires first implementing an ONNX parser for the operation. This involves registering the node type and creating an operation converter that extracts the shapes, weights, and attributes from the ONNX graph and constructs the bounded node. The interval bound computation, \CROWN relaxation, and optional handling of $\alpha$-parameters must be defined in the node class. 

\begin{table}[t!]
\centering
\begin{tabular}{llcc}
\toprule
\textbf{Category} & \textbf{Operation} & \textbf{CROWN} & \textbf{$\alpha$-CROWN} \\
\midrule
\multirow{3}{*}{Linear}
  & Fully Connected (Gemm) & \checkmark & \checkmark \\
  & Convolution            & \checkmark & \checkmark \\
  & BatchNormalization     & \checkmark & \checkmark \\
\midrule
\multirow{2}{*}{Nonlinear Activation}
  & ReLU                  & \checkmark & \checkmark \\
  & Sigmoid               & \checkmark & -- \\
\midrule
\multirow{2}{*}{Elementwise}
  & Add                   & \checkmark & \checkmark \\
  & Sub                   & \checkmark & \checkmark \\
\midrule
\multirow{3}{*}{Shape}
  & Reshape               & \checkmark & \checkmark \\
  & Flatten               & \checkmark & \checkmark \\
  & Gather / Slice        & \checkmark & \checkmark \\
\bottomrule
\end{tabular}
\caption{Currently supported operations in the \sys bound propagator.}
\label{tab:supported-ops}
\end{table}

The base Bounded Node class inherits from \code{torch::nn::Module}, the Torch base class, which provides automatic parameter tracking and device transfers required for all bound propagation procedures. Additionally, since all tensor operations within a Torch module contribute to a differentiable computation graph, gradients flow through the backward pass with respect to the learned slope parameters $\alpha$ without requiring custom differentiation logic. The Bounded Node class implements the operation-specific logic for IBP and \CROWN. 

\subsubsection{Bounding Constants and Inputs}

Bounded Constant and Input nodes are two unique node types which behave differently from other bounded nodes. Rather than transforming the affine coefficients during backward substitution, they serve as the terminal nodes in the computational graph.

The Bounded Constant Nodes contain fixed tensors loaded from the ONNX file, such as weight matrices, bias vectors, and scaling factors. Contrary to all other node types, their \code{isPerturbed()} method returns \code{false}, ending the backward traversal. During propagation, constant nodes impact only the bias term of the affine bound expression. The constant value is applied to the bias, shifting the bound by the constant value. The nodes do not impact the coefficient matrices in any way. 

The Bounded Input Nodes represent the network's input variables. Their \code{isPerturbed()} method returns \code{true}, reflecting the uncertainty in the inputs associated with the input bounds. In a similar manner to the constant nodes, the input nodes terminate the backward bound propagation process via their no-op \code{boundBackward()} method. When the backward pass reaches an input node, the accumulated affine coefficients represent the final symbolic bounds in terms of the input variables.

\subsubsection{Bounded $\alpha$ Nodes}

Activation functions that support $\alpha$-optimization also inherit from an Optimizable Node class. This intermediate class registers the node with the $\alpha$-optimization manager and provides the interface for storing and retrieving per-neuron learnable slope parameters. Separate $\alpha$ parameters are maintained for lower-bound and upper-bound optimization, as the slope that produces the tightest lower bound may differ from the one that produces the tightest upper bound. These parameters are allocated lazily during the first call to \code{boundBackward()}, since only at that point does the node know which neurons are unstable from the IBP pre-pass; stable neurons use a fixed slope and do not require learnable parameters.

\section{Analysis Engine}
\label{sec:analysis-engine}
The analysis engine implements two analysis methods. \CROWN constructs linear upper and lower bounds by propagating symbolic coefficients backward through the graph, while $\alpha$-CROWN tightens those bounds by optimizing learnable slope parameters for nonlinear activations.

\subsection{Key Implementation Decisions}
\label{sec:key-decisions}
Several aspects of a bound propagator's implementation are not prescribed by the \alphaCROWN algorithm, yet meaningfully affect the precision and efficiency of the analysis. \sys makes three such decisions deliberately:
\begin{itemize}
    \item a pending-count backward traversal, guaranteeing correctness at join points on general computational graphs (Section~\ref{sec:backward-traversal});
    \item a two-level $\alpha$ map with lazy allocation, enabling joint optimization of intermediate and output-level relaxations (Section~\ref{sec:alpha-storage});
    \item straight-through monotonic improvement of intermediate bounds during optimization (Section~\ref{sec:ste}).
\end{itemize}
These choices distinguish \sys from a direct re-implementation of \autolirpa and are key contributors to the differences in bound quality observed in Section~\ref{sec:bound_width_runtime}.

\subsection{\CROWN}
The CROWN analysis is split into three steps. First, IBP is run to obtain initial pre-activation bounds and identify unstable neurons. Second, the bounds at intermediate nodes are refined via a series of intermediate \CROWN passes. Third, a final backward pass is invoked from the output node. 

\subsubsection{Initialization}
Each backward pass begins by initializing the affine coefficient matrices $\mathcal{A}_L, \mathcal{A}_U$ and bias terms $\beta_L, \beta_U$. The coefficient matrices are initialized according to the type of backward pass: $\mathcal{I}$ for intermediate passes, where each row bounds an individual neuron, or $C$ for the final pass, where the rows encode the output specification. The bias terms are initialized to zero.

\subsubsection{Initial Interval Bound Propagation}
Interval Bound Propagation (IBP) computes a forward pass through the network using interval arithmetic. For each node $v_i$ in forward topological order, the output interval, $[\ell_i, u_i]^{\text{IBP}}$, is computed from the intervals of its input dependencies. For input nodes, the IBP bounds are simply the input bounds given in the specification. IBP bounds are computationally cheap but typically loose; however, they are required for the \CROWN backward pass to distinguish a neuron's activation status. 

The IBP bounds classify each neuron's activation status. A neuron $k$ at node $v_i$ is unstable if $\ell_i^{\text{IBP}} < 0 < u_i^{\text{IBP}}$, and stable otherwise. The linear relaxation for stable neurons is exact, so their IBP bounds are already as tight as \CROWN would produce. Consequently, \CROWN is applied only to unstable neurons, while the IBP bounds are retained for stable neurons during concretization.

In \sys, IBP bounds are stored in a per-node map and are detached from the computational graph, as they are fixed reference bounds and do not contribute to gradient-based optimization. For the input nodes, the IBP bounds are copied into the concrete bound map, as they serve as the ground truth. Non-input bounds only have their concrete bounds set in subsequent \CROWN backward passes. 

\subsubsection{Tightened Intermediate Bounds} \label{sec:crown-tighten-intermediate}
To produce the tightest possible bounds during the final pass, each nonlinear node requires tight bounds on its inputs to produce a tight linear relaxation. These bounds are commonly referred to as pre-activation bounds. Each nonlinear node's relaxations depend on the tightness of its pre-activation bounds.
When the pre-activation bounds are tighter, fewer neurons are classified as unstable, reducing the number of linear approximations. Additionally, when a nonlinear node has tighter pre-activation bounds, the area of the convex relaxation is reduced, allowing for more accurate approximations. 

\sys tightens each nonlinear node's input bounds via a series of subsequent \CROWN passes, triggered during a backward depth-first search (DFS) from the output node. For each node whose \code{requiresInputBounds} list is non-empty, \sys triggers tightened bound computation on each of the inputs in the list. For each of these passes, the tightened bound computation follows a series of hierarchical heuristic decisions improving efficiency. First, \code{checkIBPIntermediate()} walks backward from the target node: if every ancestor has its \code{ibpIntermediate} flag set (e.g., all ancestors are affine layers), IBP bounds are exact and no \CROWN pass is needed. Second, \code{isFirstLinearLayer()} checks whether the node is a linear layer whose sole dependency is the input node. In this case, IBP is exact and equally as tight as \CROWN. Third, \sys checks if the unstable set of neurons of the node is empty. If so, IBP is exact and \CROWN is skipped. When unstable neurons do exist, a CROWN Backward pass is started from the intermediate node. The resulting concrete bounds are stored in the concrete bounds map, saved within the node, and registered with the bounded model.  

The above describes standard backward \CROWN. \sys also provides support for CROWN-IBP, a variant of CROWN which skips this intermediate bound tightening. In this case, \sys computes the IBP bounds and then directly begins the final backward pass from the output node. This lowers the computational cost by skipping the intermediate \CROWN backward passes; however, this results in less precise bounds. 

\subsubsection{Backward Traversal}
\label{sec:backward-traversal}
The backward pass substitutes linear bound coefficients from a start node toward the input using a scheduled reverse traversal. A backward $BFS$ from the output node first identifies the reachable set $\mathcal{R}$, i.e. every node reachable from the output. For each node in $\mathcal{R}$, a pending count is initialized to the number of its dependents. The output node's count is initialized to zero. A node is then added to the queue $Q$ only when its count reaches zero, meaning all of its dependents have been processed and the relaxations have been accumulated into $\mathcal{A}^{(v)}$. This guarantees correctness at join points (i.e., nodes with multiple dependents such as the input to a residual addition) where $\mathcal{A}^{(v)}$ must reflect contributions from all incoming paths before the node's own backward relaxation is invoked. A simple reverse topological order does not provide this guarantee, since it fixes a static traversal order that may process a node before all its dependents have finished. 

\subsubsection{Relaxation and Bias Propagation}
At each node visited during the backward traversal, the operation-specific node transforms the incoming bound coefficients $\mathcal{A}_L^{(c)}, \mathcal{A}_U^{(c)}$ into new coefficients for each of the node's input dependencies. For linear layers, this transformation is exact. The $\mathcal{A}$ matrix is composed with the layer's weight matrix via $\mathcal{A}^{(d)} = \mathcal{A}^{(c)} W$, and the layer's bias vector $b$ is combined into the bound's bias term via $\beta^{(d)} = \beta^{(c)} + \mathcal{A}^{(c)} b$. For nonlinear layers, such as ReLU, the transformation applies a relaxation that produces new coefficients and additional bias terms. Active neurons leave the coefficient matrices unchanged with no bias contribution. Inactive neurons zero out their corresponding rows, and unstable neurons scale the coefficients by a relaxation slope, with an intercept accumulated into the bias terms.

The resulting coefficients are accumulated at each dependency via summation. At points where multiple paths converge, summing the $\mathcal{A}$ matrix represents the total linear dependence on that node; however, bias terms are attached only to the first dependency of that node. This prevents double-counting the bias. A bias term originating from a specific node should only contribute once to the final bounds. If it was propagated to each of the input dependencies, it would be summed once per-path when those paths later merge. Restricting bias propagation to the first dependency ensures tight, sound bounds.

\subsection{\alphaCROWN}
The \alphaCROWN module extends \CROWN by introducing learnable slope parameters that can tighten relaxations for non-linear activations. The \CROWN analysis module performs all bound computation, while \alphaCROWN manages $\alpha$ parameter storage and the optimization loop. At each optimization step, the full \CROWN backward pass is re-run, producing output bounds that depend on the current $\alpha$ parameters. The loss is backpropagated through the entire computation to update the $\alpha$ parameters via projected gradient descent. 

\subsubsection{Initialization from \CROWN}
Before the optimization loop begins, \sys runs a full standard CROWN analysis backward pass without gradients to establish the network's relaxation slopes. During this pass each nonlinear node records the lower-bound slope that standard \CROWN computes for every neuron via a heuristic choice. These slopes serve as the initial values for the $\alpha$ parameters.

$\alpha$ parameters are created lazily rather than pre-allocated. The first time a \CROWN backward pass visits a nonlinear node, whether during the intermediate bound tightening or final pass, \sys identifies the unstable neurons at the node using the IBP bounds, extracts the corresponding \CROWN slopes, and creates the $\alpha$ tensor. Stable neurons do not have an associated $\alpha$ value, since their transformers are exact: active neurons always use a slope of~$1$, and inactive neurons always a slope of~$0$. 

\subsubsection{Alpha Storage Structure} \label{sec:alpha-storage}
To allow for the joint optimization of output-level and intermediate-level relaxations, \sys organizes the $\alpha$ parameters in a two-level map indexed by (node, start key). The start key identifies which backward pass created the parameter. As described previously, \sys runs separate \CROWN passes to tighten pre-activation bounds, each originating from a different intermediate node. When these passes encounter nonlinear nodes, $\alpha$ parameters are created with the start key set to the node originating the backward pass. The final backward pass from the output node similarly creates $\alpha$ parameters keyed to the output node.

At the start of each optimization step, all $\alpha$ tensors are flattened across (node, start key) pairs into a single list and registered with the optimizer, so that a single gradient step updates both the intermediate and output-level parameters jointly.

\subsubsection{Intermediate Bound Tightening}
Although the $\alpha$ parameters originating from intermediate passes can be optimized independently, the loss function used during optimization is defined solely in terms of the output bounds. No term in the loss encourages tighter intermediate bounds. Instead, the gradient signal reaches the intermediate $\alpha$ parameters implicitly. The output bounds depend on the
transformer at each nonlinear node, which depends on the node's pre-activation bounds, which are themselves computed by intermediate backward passes parameterized by their own $\alpha$ parameters. This chain of dependencies is contained in the gradient graph, so the gradient of the output-bound loss propagates through the intermediate computations and reaches the intermediate $\alpha$ parameters. 

In practice, this gradient signal encourages tighter intermediate bounds. A tighter pre-activation bound at a nonlinear node may stabilize some neurons, reducing the number of unstable neurons and hence the number of relaxations needed. When the $\alpha$ parameters originating from intermediate passes are not included, the pre-activation bounds for each nonlinear layer are fixed at what the \CROWN passes produce.

An alternative implementation would add an explicit intermediate bound tightness term to the loss, directly encouraging each pre-activation to tighten. This introduces a multi-objective optimization in which the gradients for the output and intermediate terms can conflict. In this scenario, the output loss may prefer a particular intermediate $\alpha$ value which does not correspond to the tightest intermediate bound at that node. Balancing these competing objectives would require additional hyperparameters and be difficult to adjust across network architectures and benchmark instances. \sys's single objective indirectly tightens the intermediate bounds, while also being generalizable to any network.

\subsubsection{Optimization Loop}
The optimization loop iteratively refines the $\alpha$ parameters to tighten the certified bounds. Following the initial \CROWN analysis described above, the allocated $\alpha$ tensors are collected from every nonlinear node and registered with the optimizer.

Each iteration then proceeds as follows. First, the \CROWN state is cleared and a full \CROWN analysis pass is re-run, producing new bounds $\hat{\ell}_t, \hat{u}_t$ that depend on the current $\alpha$ parameters. The loss is then computed from the bound being optimized: $\mathcal{L} = -\sum_s \hat{\ell}_s$ when optimizing lower bounds, or $\mathcal{L} = \sum_s \hat{u}_s$ when optimizing upper bounds. By minimizing this loss, we tighten the target bounds. The optimizer then performs a gradient step, and the $\alpha$ parameters are projected back onto $[0,1]$ via clamping. The learning rate is decayed after each iteration. 

\subsubsection{Monotonic Improvement via Straight-Through Estimation} \label{sec:ste}
\sys's indirect optimization can degrade the precision of intermediate abstract values. The intermediate $\alpha$ parameters are optimized against the output loss; if loosening an intermediate bound reduces the output loss, the optimizer is free to do so. These looser bounds can cascade: more neurons become unstable, more relaxations are needed, and overall precision degrades. To prevent this, \sys enforces monotonic improvement of the intermediate bounds using a straight-through estimator (STE), in the sense that the clamped bound is used in the forward computation while gradients bypass the clamp.

At the start of \alphaCROWN optimization, reference bounds $[\ell_i^{\mathrm{ref}}, u_i^{\mathrm{ref}}]$ are captured from the initial \CROWN pass for each intermediate node~$i$. In subsequent iterations, each computed lower bound $\hat{\ell}_i$ is clamped against the reference:
\begin{equation}
  \ell_i^{\mathrm{STE}} \;=\; \max\bigl(\ell_i^{\mathrm{ref}},\; \hat{\ell}_i\bigr).\texttt{detach()} \;-\; \hat{\ell}_i.\texttt{detach()} \;+\; \hat{\ell}_i, \label{eq:ste}
\end{equation}
with the upper bound $u_i^{\mathrm{STE}}$ handled symmetrically via $\min$. The forward value of \eqref{eq:ste} is $\max(\ell_i^{\mathrm{ref}}, \hat{\ell}_i)$, so subsequent \CROWN backward passes always use the tighter of the reference and computed bounds, and relaxations never worsen. Because the first two terms are detached, the gradient flows only through $\hat{\ell}_i$, as if the clamping did not occur. Without this, whenever the clamp is active the intermediate $\alpha$ parameters would receive zero gradient and could never recover from a temporarily loose bound. Finally, the reference bounds are updated element-wise whenever an iteration produces a tighter intermediate bound, so the bounds used by intermediate \CROWN passes are guaranteed to be at least as tight as the best seen thus far.

\subsubsection{Separate Lower and Upper Optimization}
\sys optimizes the lower and upper output bounds in separate passes, each maintaining its own set of $\alpha$ parameters. In the standard triangle relaxation of each unstable ReLU, the upper bound is fixed (the line connecting the endpoints of the pre-activation interval), while the lower bound is parameterized by a free slope $\alpha \in [0,1]$. These slopes are the only learnable parameters in both passes. The lower-bound loss $-\sum \hat{\ell}$ drives each $\alpha$ toward the slope that maximizes the concrete lower bound; conversely, the upper-bound loss $\sum \hat{u}$ drives the same $\alpha$ toward the slope that minimizes the concrete upper bound. Because gradients flow differently through the bound computation under each objective, the optimal $\alpha$ values typically differ between the two passes. When upper and lower bounds share a single $\alpha$, the resulting output bounds are often less precise than when each pass uses its own. The dedicated two-pass approach adds computational cost, but allows every $\alpha$ to converge to the value best suited for that bound direction.

An important distinction between the two optimization passes is the goal of the $\alpha$ parameters. During both the upper and lower optimization passes, the $\alpha$ parameters represent the slope of the lower bound of the relaxation. For ReLU relaxations, the upper bound of the relaxation is fixed (the secant from $(\ell, 0)$ to $(u, u)$, with slope $\tfrac{u}{u - \ell}$) and is not optimized. The two passes therefore adjust the same parameter against different objectives: the lower-bound pass tunes $\alpha$ to maximize the concrete output lower bound, while the upper-bound pass tunes the same $\alpha$ to minimize the concrete output upper bound. Aside from the loss, the two passes are identical.

\subsection{Front End}
\sys currently provides a Command-Line Interface (CLI), a native C++ API, and a Python API. The CLI follows the VNN-COMP conventions, making \sys usable with existing benchmarking pipelines. The C++ API allows \sys to be linked directly into verifiers written in C++, such as Marabou, without interpreter start-up or foreign-function overhead. The Python API exposes the same underlying backend to users who wish to script experiments, embed \sys in a larger tool, or specify bounds that do not fit the VNN-LIB format. 

The CLI expects an ONNX network and a VNN-LIB specification as input. By default, \sys runs standard backward \CROWN and returns the resulting bounds. All other options are optional flags. Figure~\ref{fig:cli-example} shows an example of CLI usage.

\begin{figure}[!htbp]
\begin{tcolorbox}[title=CLI Example, colback=gray!5,
                  colframe=black!60, fonttitle=\bfseries]
\begin{minted}{bash}
# Run verification with alpha-CROWN, optimizing both bounds
luna --input path/to/model.onnx       \   # ONNX network
      --vnnlib path/to/property.vnnlib \  # VNN-LIB specification
      --method alpha-crown             \  # analysis method
      --optimize-lower                 \  # optimize lower bounds
      --optimize-upper                 \  # optimize upper bounds
      --lr 0.5                         \  # learning rate
      --iterations 20                     # number of iterations
\end{minted}
\end{tcolorbox}
\caption{Example CLI usage: verifying an ONNX model against a VNN-LIB
         property using $\alpha$-CROWN with optimized upper and lower bounds.}
\label{fig:cli-example}
\end{figure}

The C++ API is the primary route for embedding \sys into a complete verifier. A client constructs a \code{TorchModel} from an ONNX file, sets the input bounds, and invokes the desired analysis directly, with no interpreter involved. Figure~\ref{fig:cpp-api-example} shows a minimal end-to-end use of the C++ API.

\begin{figure}[!htbp]
\begin{tcolorbox}[title=C++ API Example, colback=gray!5,
                  colframe=black!60, fonttitle=\bfseries]
\begin{minted}{cpp}
#include "engine/TorchModel.h"

using namespace NLR;

// Load the ONNX model
TorchModel model("path/to/model.onnx");

// Define input bounds
unsigned n = model.getInputSize();
torch::Tensor lb = torch::full({n}, -1.0);
torch::Tensor ub = torch::full({n},  1.0);
model.setInputBounds(BoundedTensor<torch::Tensor>(lb, ub));

// Compute output bounds using alpha-CROWN
BoundedTensor<torch::Tensor> result =
    model.runAlphaCROWN(/*optimizeLower=*/true,
                        /*optimizeUpper=*/true);
\end{minted}
\end{tcolorbox}
\caption{Example C++ API usage: loading an ONNX model, setting input
         bounds, and computing optimized bounds via $\alpha$-CROWN.}
\label{fig:cpp-api-example}
\end{figure}

The Python API offers finer-grained control over the verification process. Users load an ONNX model through the \code{TorchModel} class, specify input bounds as NumPy arrays, and invoke \code{compute\_bounds} with a chosen analysis method. Figure~\ref{fig:python-api-example} illustrates a minimal end-to-end use of the API.

All three interfaces expose the same set of configuration options for the analysis passes, including the time limit, the analysis method (\CROWN, or \alphaCROWN), which bounds to optimize (upper, lower, or both), and the number of iterations used in \alphaCROWN. 

\begin{figure}[!htbp]
\begin{tcolorbox}[title=Python API Example, colback=gray!5,
                  colframe=black!60, fonttitle=\bfseries]
\begin{minted}{python}
from lirpapy import TorchModel
import numpy as np

# Load the ONNX model
onnx_path = "path/to/model.onnx"
model = TorchModel(onnx_path)

# Define input bounds
input_size = model.getInputSize()
lower_bounds = np.full(input_size, -1.0)
upper_bounds = np.full(input_size,  1.0)
model.setInputBounds(lower_bounds, upper_bounds)

# Compute output bounds using alpha-CROWN
result = model.compute_bounds(method="alphaCROWN")
\end{minted}
\end{tcolorbox}
\caption{Example Python API usage: loading an ONNX model, setting input
         bounds, and computing an optimized bound via $\alpha$-CROWN.}
\label{fig:python-api-example}
\end{figure}

\subsection{Code, Installation, and Availability}
\sys is implemented in 20k lines of C++17 code with Python bindings via pybind11. The build system uses CMake and automatically downloads dependencies (libtorch, Protobuf, ONNX, pybind11) when not found locally. The project includes a comprehensive test suite of over 45 test files organized into unit tests (covering individual node types, parsers, and analysis routines), integration tests (end-to-end verification pipelines), and property-based tests (e.g., validating that \CROWN bounds are at least as tight as IBP bounds on every test instance). The \sys source code is available on GitHub\footnote{\href{https://github.com/ai-ar-research/luna}{https://github.com/ai-ar-research/luna}} under the permissive modified BSD open-source license.
\section{Evaluation}

\subsection{Experimental Setup}
We compare against \autolirpa~\cite{xu2020automatic}, which underlies the top two performers, $\alpha$-$\beta$-CROWN and NeuralSAT, in the past two VNN-COMPs~\cite{brix2024fifth,kaulen2025}. We consider regular track benchmarks from VNN-COMP 2025~\citep{kaulen2025}. \sys currently supports 11 of the 16 benchmark sets. The networks range from smaller controller networks with thousands of parameters and two-dimensional inputs to larger convolutional benchmarks with millions of parameters and complex multi-dimensional inputs, covering a broad spectrum of use cases, architectures, and scales. Benchmark properties are shown in Table~\ref{tab:benchmarks}. Full per-instance results for all the reported evaluations are publicly available.\footnote{\url{https://drive.google.com/drive/folders/1m03QxkoQ1OlPex7qrs6lwP26XEzuDlR0?usp=share_link}}

\begin{table*}[h!]
\centering
\setlength{\abovecaptionskip}{10pt}
\resizebox{\textwidth}{!}{%
\begin{tabular}{lllllr}
\toprule
Category & Benchmark & Application & Network Types & \# Params & Effective Input Dim \\
\midrule
Complex
  & LinearizeNN       & NN controller approximation             & FC. + Conv. + Vision Transformer + Residual + ReLU      & 203,000              & 4         \\
\midrule
\multirow{5}{*}{\shortstack{CNN \\ \& ResNet}}
  & Collins RUL CNN   & Condition Based Maintenance             & Conv. + ReLU, Dropout                                   & 60,000 -- 262,000    & 400 -- 800 \\
  & cifar100          & Image Classification                    & FC + Conv. + Residual, ReLU + BatchNorm                 & 2.5M -- 3.8M         & 3072      \\
  & Metaroom          & ---                                     & Conv. + FC, ReLU                                        & 466,000 -- 7.4M      & 5376      \\
  & MalBeWare         & Malware Classification                  & FC + Conv. + ReLU                                       & 102,000 -- 1.5M      & 4,000     \\
  & cersyve           & Neural Certificates for Control Systems & FC + Residual + ReLU                                    & 1,000 -- 4,000       & 2 -- 6    \\
\midrule
\multirow{5}{*}{FC}
  & TLL Verify Bench  & Two-Level Lattice NN                    & Two-Level Lattice NN (FC. + ReLU)                       & 17,000 -- 67M        & 2         \\
  & Acas XU           & Collision Detection                     & FC. + ReLU                                              & 13,000               & 5         \\
  & safeNLP           & Sentence classification                 & FC. + ReLU                                              & 4,000                & 30        \\
  & CORA              & Image Classification                    & FC. + ReLU                                              & 575,000, 1.1M        & 784, 3072 \\
  & SAT ReLU          & Verifier Test                           & FC. + ReLU                                              & 100 -- 35,000        & 2 -- 100  \\
\bottomrule
\end{tabular}%
}
\caption{Considered benchmarks from the VNN-COMP benchmark suite. \# Params refers to the number of learnable parameters (weights and biases) in the network. Effective Input Dim is the number of input dimensions over which input properties are defined (e.g., cifar100 processes $32\times32\times3$ images, giving an effective input dimension of 3,072).}
\label{tab:benchmarks}
\end{table*}

Key hyperparameters for the $\alpha$-CROWN analysis include the number of optimization iterations and the learning rate. We fix those two parameters for both \sys and \autolirpa to 20 and 0.5, respectively. Aside from these shared hyperparameters, we used each tool's default configuration for optimizing upper and lower bounds.

\subsection{Results and Analysis}

\subsubsection{Comparison on CPU}
\label{sec:bound_width_runtime}
In the following section we report three metrics: average bound width, average runtime in seconds, and the number of finished instances. An instance is counted as finished if the full analysis terminates before the 300\,s wall-clock timeout and returns concrete output bounds together with a verdict of either verified or unknown; unfinished instances are timeouts. Finished therefore measures whether the analysis completed, not whether the property was decided. We first compare the performance of the two tools on CPU, using a cluster equipped with Dell PowerEdge R6525 CPU Servers. Each job is given 40 CPU cores, a 192\,GB memory limit, and a 300\,s wall-clock time limit.

\begin{table}[!t]
\centering
\setlength{\abovecaptionskip}{10pt}
\begin{tabular}{lcccccc}
\toprule
\multirow{2}{*}{Bench. (\# inst.)}
  & \multicolumn{3}{c}{\sys}
  & \multicolumn{3}{c}{\autolirpa} \\
\cmidrule(lr){2-4}\cmidrule(lr){5-7}
  & {Width} & {Time (s)} & {Finished}
  & {Width} & {Time (s)} & {Finished} \\
\midrule
acasxu\_2023 (184)                    &    \textbf{867.42} &   \textbf{1.41} & 184 &    882.90 &   6.19 & 184 \\
cersyve (12)                          &      \textbf{4.27} &  \textbf{1.23} &  12 &      4.38 &   6.77 &  12 \\
cifar100\_2024 (200)                  &      4.63 &  44.88 & \textbf{170} &      \textbf{4.62} & 136.59 & 100 \\
collins\_rul\_cnn\_2022 (62)          &     \textbf{35.80} &   \textbf{3.57} &  62 &     35.98 &   5.78 &  62 \\
cora\_2024 (60)                       &  12073.43 &   \textbf{8.37} &  60 &  \textbf{12067.97} &   9.53 &  60 \\
linearizenn\_2024 (60)                &     \textbf{43.44} &   \textbf{2.96} &  60 &     48.53 &   6.59 &  60 \\
malbeware (150)                       &      5.42 &   \textbf{4.26} & 150 &      5.42 &   5.30 & 150 \\
metaroom\_2023 (100)                  &      0.16 &   \textbf{1.68} & \textbf{100} &      0.16 &   5.30 &  95 \\
safenlp\_2024 (1046)                  &     18.82 &   \textbf{0.47} & 1046 &     18.82 &   5.10 & 1046 \\
sat\_relu (98)                        &     58.99 &   \textbf{0.51} &  98 &     58.99 &   5.13 &  98 \\
tllverifybench\_2023 (32)             &    \textbf{207.98} &  \textbf{18.63} &  \textbf{32} &    213.18 &  45.03 &  20 \\
\bottomrule
\end{tabular}
\caption{CPU comparison of \sys and \autolirpa. Width: average interval across output dimensions. Time: average runtime per instance. Finished: instances for which the analysis terminated and returned bounds within the 300\,s timeout. Width and Time averages include only instances where both tools finished.}
\vspace{-0.5cm}
\label{tab:bound_width_runtime}
\end{table}

Table~\ref{tab:bound_width_runtime} shows the average bound width and runtime results of \sys and \autolirpa on the evaluated benchmarks across all instances and properties. \sys achieves a faster runtime and finishes at least as many instances as \autolirpa across all 11 benchmarks. In terms of bound quality, \sys is competitive with \autolirpa, computing tighter average bounds on $5$ of $11$ benchmarks, matching bounds on $4$ and producing slightly looser bounds on $2$. The two benchmarks where \autolirpa produces slightly tighter bounds, cifar100\_2024 ($0.22\%$ wider) and cora\_2024 ($0.05\%$ wider), are paired with substantial efficiency gains for \sys. The most notable efficiency gain is cifar100\_2024, where \sys completes 170 of 200 instances, while running over $3\times$ faster. Across the full suite of evaluated benchmarks, \sys completes $1,974$ of $2,004$ total instances ($98.5\%$), compared to \autolirpa's $1,887$ of $2,004$ total instances ($94.2\%$).

% \changed{
% Since \alphaCROWN optimizes the relaxation slopes by projected gradient descent over a nonconvex objective, the resulting tightness depends on implementation choices the algorithm does explicitly enumerate, even under matched budgets. A primary example among these is how often intermediate pre-activation bounds are refreshed: \sys recomputes them throughout optimization and optimizes intermediate and output $\alpha$ parameters jointly under the monotonic-improvement clamp of Section~\ref{sec:ste}, so a bound that tightens mid-optimization can stabilize neurons and shrink relaxations downstream. Differences in the order of operations during relaxation and back-substitution, and in how gradients accumulate through the computation graph, further cause the two tools to converge to different local optima. This accounts for the cases where \sys is tighter, such as linearizenn\_2024, as well as the small, unsystematic differences elsewhere.}

\begin{figure}[!h]
    \centering
    \includegraphics[width=0.34\linewidth]{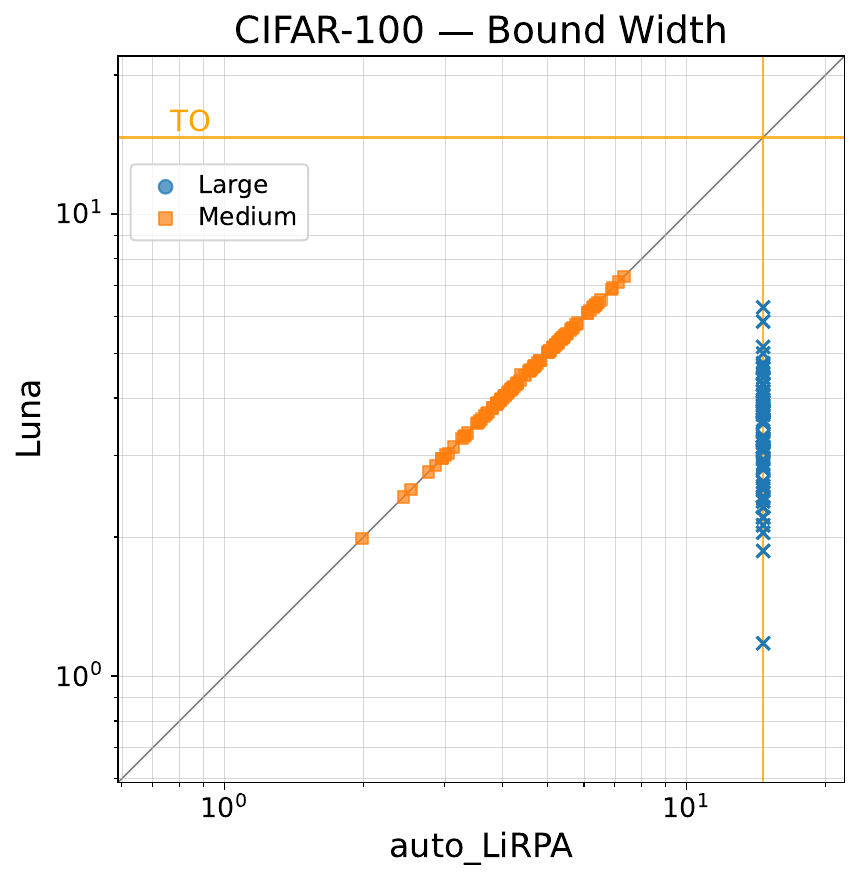}
    \hfil
    \includegraphics[width=0.34\linewidth]{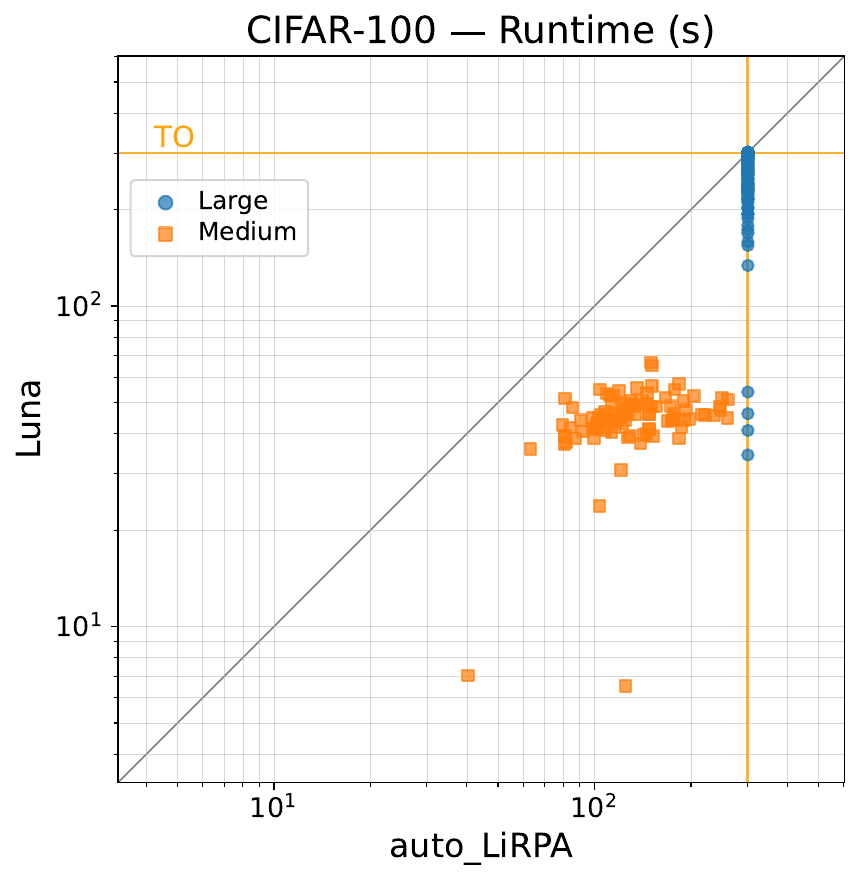}
    \caption{Bound width (left) and runtime (right) for Cifar100 2024}
    \label{fig:cifar100_2024}
\end{figure}

Figure~\ref{fig:cifar100_2024} shows per-instance and per-property results for the two tools on the cifar100\_2024 benchmarks. The benchmark contains Medium and Large convolutional networks with 2.5M--3.8M parameters. In the bound-width plot (\Cref{fig:cifar100_2024}, left), both \sys and \autolirpa compute nearly identical bounds on all Medium network instances; however, \autolirpa times out under the 300\,s limit on all of the Large network instances. Both tools ran in their default configuration, using matrix mode with heuristic switching to patches. The runtime plot (\Cref{fig:cifar100_2024}, right) shows a similar pattern, with \sys finishing bound computation consistently faster than \autolirpa on the Medium instances, as well as finishing on the majority of the Large instances. Since the following GPU evaluation eliminates these timeouts, with both tools completing every instance under the same configuration and \sys remaining consistently faster, the GPU results provide a cleaner basis for the efficiency conclusions. %While the aggregate statistics in Table~\ref{tab:bound_width_runtime} summarize overall trends, the scatter plot shows finer-grained patterns dependent on network size and property type that are not visible in the averages. 
Scatter plots for all CPU results can be found in Appendix~\ref{app:per-instance-CPU-plots}.
 
%The cifar100\_2024 plot (Figure~\ref{fig:cifar100_2024}) shows a significant result regarding \sys's performance on larger networks. 
 
\subsubsection{Comparison on GPU}
\label{sec:gpu-comparison}
%The previous results evaluated both tools on CPU-only hardware. However, state-of-the-art neural network verifiers increasingly leverage GPU acceleration to exploit the inherent parallelism in these analyses, particularly in the matrix multiplications underlying the repeated backward passes.
To assess how \sys and \autolirpa compare when GPU resources are available, we repeat the preceding CPU evaluation on GPU-equipped hardware using identical hyperparameters. All experiments in this section were run on a cluster equipped with NVIDIA DGX A100 GPU servers. Each job is given 1 NVIDIA A100 GPU (80 GB GPU memory), 40 CPU cores, a 64 GB host-memory limit, and a 300\,s wall-clock time limit.

\begin{table}[!t]
\centering
\setlength{\abovecaptionskip}{10pt}
\begin{tabular}{lcccccc}
\toprule
\multirow{2}{*}{Bench. (\# inst.)}
  & \multicolumn{3}{c}{\sys}
  & \multicolumn{3}{c}{\autolirpa} \\
\cmidrule(lr){2-4}\cmidrule(lr){5-7}
  & {Width} & {Time (s)} & {Finished}
  & {Width} & {Time (s)} & {Finished} \\
\midrule
acasxu\_2023 (185)                    & \textbf{863.74} & \textbf{3.94} & 185 &     879.27 &  13.06 & 185 \\
cersyve (12)                          & \textbf{4.27} & \textbf{3.79} &  12 &       4.38 &  13.08 &  12 \\
cifar100\_2024 (200)                  &       4.23 & \textbf{13.32} & 200 &       4.23 &  25.37 & 200 \\
collins\_rul\_cnn\_2022 (62)          & \textbf{35.80} & \textbf{2.45} &  62 &      35.98 &   9.33 &  62 \\
cora\_2024 (60)                       &   12073.44 & \textbf{5.25} &  60 & \textbf{12067.99} &  11.93 &  60 \\
linearizenn\_2024 (60)                & \textbf{43.44} & \textbf{4.95} &  60 &      48.35 &  11.92 &  60 \\
malbeware (150)                       &       5.42 & \textbf{2.80} & 150 &       5.42 &   7.70 & 150 \\
metaroom\_2023 (100)                  & \textbf{12.78} & \textbf{3.11} & 100 &      13.17 &   9.07 & 100 \\
safenlp\_2024 (1046)                  &      18.82 & \textbf{2.27} & 1046 &      18.82 &   8.38 & 1046 \\
sat\_relu (98)                        &      58.99 & \textbf{2.67} &  98 &      58.99 &   8.92 &  98 \\
tllverifybench\_2023 (32)             & \textbf{293.12} & \textbf{14.45} &  32 &     303.24 &  21.22 &  32 \\
\bottomrule
\end{tabular}
\caption{GPU comparison of \sys and \autolirpa. Width: average interval across output dimensions. Time: average runtime per instance. Finished: instances for which the analysis terminated and returned bounds within the 300\,s timeout. Width and Time averages include only instances where both tools finished.}
\vspace{-0.5cm}
\label{tab:gpu_shared_benchmarks}
\end{table}

Table~\ref{tab:gpu_shared_benchmarks} reports the results for the GPU-based experiments. With GPU acceleration, both tools complete all instances across every benchmark, eliminating the completion gap seen on the CPU. \sys remains consistently faster, achieving roughly a 2-3$\times$ speedup over \autolirpa on every benchmark. 

The largest gains from the GPU are on collins\_rul\_cnn\_2022, where \sys averages 2.45s compared to \autolirpa's 9.33s, and safenlp\_2024 (2.27s vs.\ 8.38s). The smallest relative gap is on tllverifybench\_2023 (14.45s vs.\ 21.22s). Here, the runtime is dominated by the large matrix multiplications required to bound tllverifybench\_2023's larger and wider network architecture, which both tools benefit substantially from GPU acceleration. 

\begin{figure}[!h]
    \centering
    \includegraphics[width=0.33\linewidth]{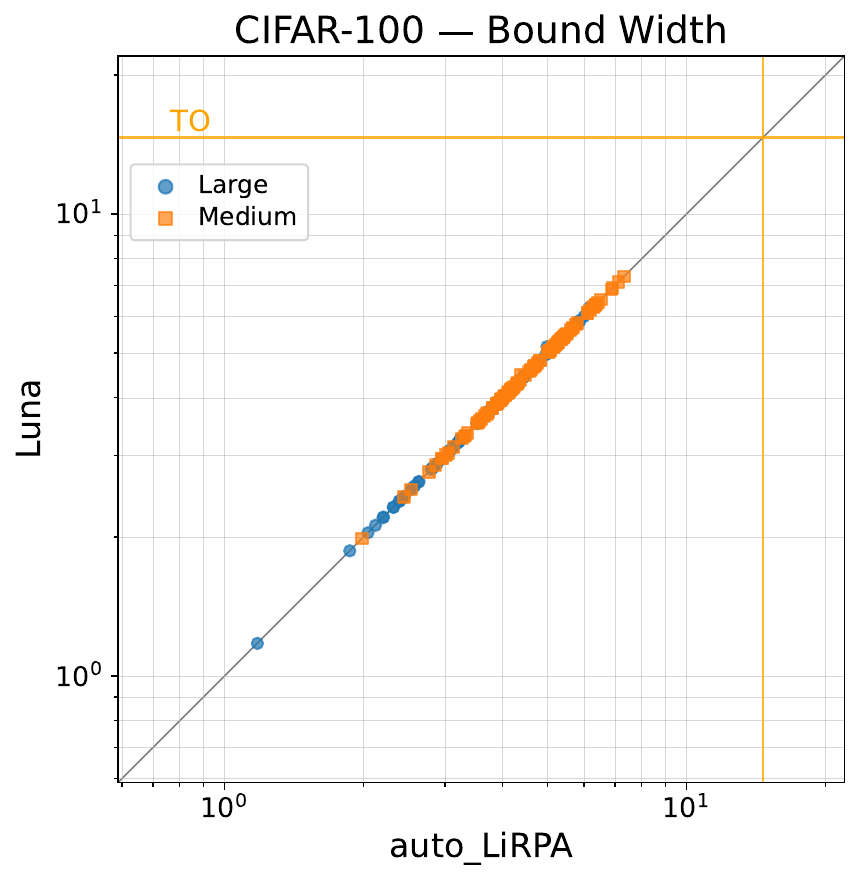}
    \hfil
    \includegraphics[width=0.33\linewidth]{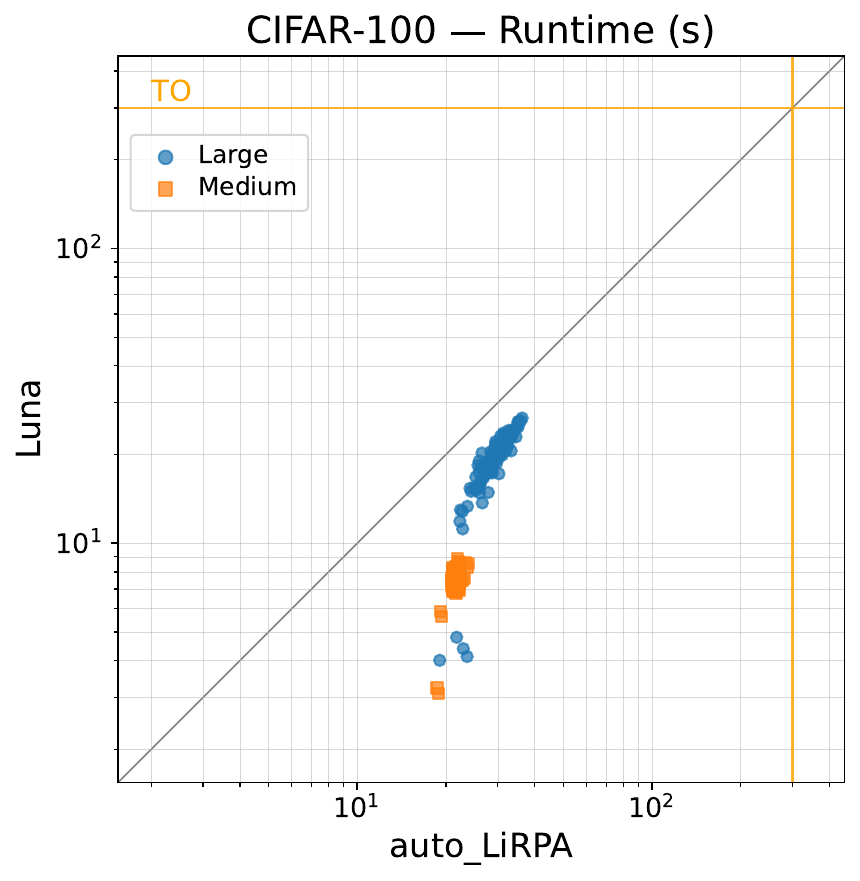}
    \caption{Bound width (left) and runtime (right) for Cifar100 2024 (GPU)}
    \label{fig:gpu_cifar100_2024}
\end{figure}

Figure~\ref{fig:gpu_cifar100_2024} shows per-instance plots for cifar100\_2024's GPU results. The plot reveals a very similar pattern in bound quality, with both \sys and \autolirpa computing competitive bound widths. For runtime, \sys consistently outperforms \autolirpa, completing both the Large and Medium benchmarks in less time. Scatter plots for all GPU results can be found in Appendix~\ref{app:per-instance-GPU-plots}. 

\subsubsection{Startup Evaluation}

%Startup time measures the overhead incurred before any analysis begins. This cost usually encompasses model parsing, graph construction, and framework initialization. 

One motivation for \sys is reducing the integration and startup cost that is inherent to Python-based tools. To isolate startup cost from analysis time, we measure total execution time on a pair of small networks where analysis time is negligible. This ensures the total runtime of each tool is dominated by the startup overhead. We evaluate \sys and \autolirpa on two networks, Nano and Tiny. %Appendix~\ref{app:startup-time-networks} contains specific per-network values of the Nano and Tiny benchmarks.

\begin{table}[!ht]
\centering
\setlength{\abovecaptionskip}{10pt}
\begin{tabular}{lcccc}
\toprule
\multirow{2}{*}{Network}
  & \multicolumn{2}{c}{\sys}
  & \multicolumn{2}{c}{\autolirpa} \\
\cmidrule(lr){2-3}\cmidrule(lr){4-5}
  & {Avg. Time (s)} & {Std. Dev.}
  & {Avg. Time (s)} & {Std. Dev.} \\
\midrule
Nano        &   0.53 &   0.02 &   4.81 &   0.10 \\
Tiny        &   0.53 &   0.02 &   4.78 &   0.08 \\
\bottomrule
\end{tabular}
\caption{Startup time comparison of \sys and \autolirpa (averaged over 50 iterations).}
\label{tab:startup_times}
\end{table}

Table~\ref{tab:startup_times} shows the startup time results for both \sys and \autolirpa. \sys consistently starts in 0.53 seconds for both networks, while \autolirpa requires approximately 4.80 seconds. This represents about a 9$\times$ reduction in startup overhead. Specifically, \autolirpa's overhead of 4.81s for the Nano network and 4.78s for the Tiny network is consistent with the expected additional cost of the Python interpreter and PyTorch initialization. Conversely, \sys's C++ implementation avoids this overhead, allowing startup time to be determined by parsing and graph construction. 

\section{Conclusion and Next Steps}
\label{chap:conclusion}
We presented \sys, an abstract-interpretation-based bound propagator for deep neural networks implemented in C++. \sys supports three analyses, including the popular \alphaCROWN analysis. Our evaluation showed that \sys is competitive with \autolirpa, the state-of-the-art \alphaCROWN implementation, in terms of precision, and overall more efficient on benchmarks from VNN-COMP'25.
Future work includes supporting additional ONNX operations and activations, such as bilinear and softmax functions for transformer architectures. %, and batching verification queries to eliminate redundant graph construction across specifications that share a network. 
While \sys functions readily as a standalone incomplete verifier, integrating \sys as an in-processing engine in a complete verifier such as Marabou is ongoing work. It is also important to extend \sys with proof production capability compatible with proof production in complete verifiers~\cite{elboher2025abstraction,isac2022neural,isac2026proof,duong2026generating}. % would validate its effectiveness and demonstrate the value of a native C++ abstract interpreter in a production pipeline. 
Finally, employing online learning techniques~\cite{wu2026cubing,wu2023lightweight,wilson2025per} to optimize key parameters in \sys can further improve its performance. 

\newpage
\appendix
\renewcommand{\thesection}{Appendix \arabic{section}}
\renewcommand{\thesubsection}{\arabic{section}.\arabic{subsection}}
\renewcommand{\thesubsubsection}{\arabic{section}.\arabic{subsection}.\arabic{subsubsection}}
\section{Preliminaries on Abstract Interpretation}

\subsection{Neural Networks}
\label{app:nn}

Neural networks are functions $N: \mathbb{R}^n \to \mathbb{R}^m$. In this appendix, we focus on feedforward neural networks. A feedforward neural network is given by $l$ layers $f_1 : \mathbb{R}^n \to \mathbb{R}^{n_1}, \ldots, f_l : \mathbb{R}^{n_{l-1}} \to \mathbb{R}^m$. Each layer $f_i$ is either an affine transformation or an activation function. An affine transformation is a function of the form $f_i(x) = Ax + b$ for some $A \in \mathbb{R}^{n_i \times n_{i-1}}$ and $b \in \mathbb{R}^{n_i}$. Each row of $A$ corresponds to the weight vector of an individual neuron. An activation function is applied element-wise to a layer's output, introducing non-linearity and allowing the network to learn complex patterns. In later sections, we discuss the linear relaxations for two common activation functions: the Rectified Linear Unit (ReLU), given by $f(x) = \max(0, x)$, and the sigmoid function, given by $\sigma(x) = \frac{e^x}{e^x + 1}$.

For simplicity, the following techniques are presented for feedforward networks; however, \sys can handle neural networks that can be viewed as directed acyclic computational graphs (DAGs), including those with residual connections.

%\subssection{Neurons and Layers}
%An individual component of a layer is called a neuron. There are three types of neurons: $n$ input neurons, $m$ output neurons, and hidden neurons. The value of hidden neurons are not observed. 

\subsection{Linear Relaxations}
\label{app:lirpa}

\subsubsection{ReLU Relaxation} 
\label{app:relu_relaxation}

\begin{figure}[h]
  \centering
  \begin{subfigure}[b]{0.48\textwidth}
    \centering
    \begin{tikzpicture}
    \begin{axis}[
        width=\textwidth,
        yticklabel=\empty,
        xticklabel=\empty,
        axis lines=middle,
        axis line style={latex-latex},
        axis on top=true, 
        xmin=-4, ymin=-3, xmax=5, ymax=4.5,
        samples=100,
        clip=false,
        xlabel={\(x\)},
    ]
    \addplot[fill=blue!15, draw=none, forget plot]
        coordinates {(-2,-2) (-2,0) (3,3)};
    \addplot[black, very thick, samples at={-2, 0, 3}] {max(0,x)};
    \addplot[blue, thick, dashed, domain=-2:3] {(3/5)*(x+2)};
    \addplot[red, thick, dashed, domain=-2:3] {x};
    \addplot[gray, thin, dashed] coordinates {(-2,-2.4) (-2,0)};
    \addplot[gray, thin, dashed] coordinates {(3,-2.4) (3,3)};
    \node[below] at (axis cs:-2,-2.4) {$l$};
    \node[below] at (axis cs: 3,-2.4) {$u$};
    \end{axis}
    \end{tikzpicture}
    \caption{Case 1: $u \geq |l|$, $a = 1$}
  \end{subfigure}
  \hfill
  \begin{subfigure}[b]{0.48\textwidth}
    \centering
    \begin{tikzpicture}
    \begin{axis}[
        width=\textwidth,
        yticklabel=\empty,
        xticklabel=\empty,
        axis lines=middle,
        axis line style={latex-latex},
        axis on top=true, 
        xmin=-5, ymin=-1.6, xmax=4, ymax=4.5,
        samples=100,
        clip=false,
        xlabel={\(x\)},
    ]
    \addplot[fill=blue!15, draw=none, forget plot]
        coordinates {(-3,0) (2,0) (2,2) (-3,0)};
    \addplot[black, very thick, samples at={-3, 0, 2}] {max(0,x)};
    \addplot[blue, thick, dashed, domain=-3:2] {(2/5)*(x+3)};
    \addplot[red, thick, dashed, domain=-3:2] {0};
    \addplot[gray, thin, dashed] coordinates {(-3,-1.1) (-3,0)};
    \addplot[gray, thin, dashed] coordinates {(2,-1.1) (2,2)};
    \node[below] at (axis cs:-3,-1.1) {$l$};
    \node[below] at (axis cs: 2,-1.1) {$u$};
    \end{axis}
    \end{tikzpicture}
    \caption{Case 2: $u < |l|$, $a = 0$}
  \end{subfigure}
  \caption{Linear relaxation bounds for ReLU over the unstable interval $[\ell,u]$. The upper bound is the line from $(\ell,0)$ to $(u,u)$. The lower bound is chosen as $h_L = x$ when $u \ge |\ell|$ (Case 1), and $h_L = 0$ when $u < |\ell|$ (Case 2).}
  \label{fig:relu}
\end{figure}

For a non-linear ReLU neuron with pre-activation value $z$ and pre-activation bounds $\ell \le z \le u$, the ReLU function $\sigma(z) = \max(0, z)$ is piecewise linear and falls into one of three cases.  When $u \le 0$, the neuron is inactive and $\sigma(z) = 0$. Exact linear bounds can be found by setting $\hat{\sigma}_L = \hat{\sigma}_U = 0$. Similarly, when $\ell \ge 0$, the neuron is active and $\sigma(z) = z$. The exact linear bounds are $\hat{\sigma}_L = \hat{\sigma}_U = z$.

When $\ell < 0 < u$, a neuron is unstable. Here \CROWN adopts a linear relaxation: the upper bound is the line connecting $(\ell,0)$ and $(u,u)$ given by,
\begin{equation}
    \hat{\sigma}_U(z) = \frac{u}{u - \ell}(z - \ell),
    \label{eq:reluRelax}
\end{equation}
while the lower bound becomes $\hat{\sigma}_L = \alpha\cdot z$ where $\alpha$ can be any value between 0 and 1. In practice, the choice of this value can result in different output bounds. Standard \CROWN implementations select $\alpha$ via a fixed heuristic based on the pre-activation bounds. When $u \ge |\ell|$, $\alpha = 1$, making $\hat{\sigma}_L = z$. When $u < |\ell|$, $\alpha = 0$, making $\hat{\sigma}_L = 0$.

\subsubsection{Sigmoid Relaxation}
\label{app:sigmoid_relaxation}

\begin{figure}[h]
  \centering
  % ── Case 1: stable positive (0 ≤ l ≤ u, concave region) ──────────────────
  \begin{subfigure}[b]{0.32\textwidth}
    \centering
    \begin{tikzpicture}
    \begin{axis}[
        width=\textwidth,
        yticklabel=\empty, xticklabel=\empty,
        axis lines=middle, axis line style={latex-latex},
        axis on top=true,
        xmin=-0.5, ymin=-0.2, xmax=4.5, ymax=1.2,
        samples=100, clip=false,
        xlabel={\(x\)},
    ]
    % Shaded region between h_L (chord) and h_U (tangent)
    \addplot[fill=blue!15, draw=none, forget plot]
        coordinates {(1,0.7311) (3,0.9526) (3,0.9858) (1,0.7758)};
    % Sigmoid (full range for context, thick only on [l,u])
    \addplot[black!30, thin, domain=0.5:1] {1/(1+exp(-x))};
    \addplot[black, very thick, domain=1:3]  {1/(1+exp(-x))};
    \addplot[black!30, thin, domain=3:3.5]   {1/(1+exp(-x))};
    % Upper bound: tangent at d=2 (concave => tangent lies above)
    \addplot[blue, thick, dashed, domain=1:3] {0.1050*x + 0.6708};
    % Lower bound: chord through (l,sigma(l)) and (u,sigma(u))
    \addplot[red,  thick, dashed, domain=1:3] {0.1108*x + 0.6203};
    % Drop lines
    \addplot[gray, thin, dashed] coordinates {(1,-0.13) (1,0.7311)};
    \addplot[gray, thin, dashed] coordinates {(3,-0.13) (3,0.9526)};
    \node[below] at (axis cs:1,-0.13) {$l$};
    \node[below] at (axis cs:3,-0.13) {$u$};
    \end{axis}
    \end{tikzpicture}
    \caption{$0\!\le\!\ell$: $h_U$ tangent, $h_L$ chord}
  \end{subfigure}
  \hfill
  % ── Case 2: stable negative (l ≤ u ≤ 0, convex region) ──────────────────
  \begin{subfigure}[b]{0.32\textwidth}
    \centering
    \begin{tikzpicture}
    \begin{axis}[
        width=\textwidth,
        yticklabel=\empty, xticklabel=\empty,
        axis lines=middle, axis line style={latex-latex},
        axis on top=true,
        xmin=-4.5, ymin=-0.2, xmax=0.5, ymax=0.6,
        samples=100, clip=false,
        xlabel={\(x\)},
    ]
    % Shaded region between h_L (tangent) and h_U (chord)
    \addplot[fill=blue!15, draw=none, forget plot]
        coordinates {(-3,0.0142) (-1,0.2242) (-1,0.2689) (-3,0.0474)};
    % Sigmoid
    \addplot[black!30, thin, domain=-3.5:-3] {1/(1+exp(-x))};
    \addplot[black, very thick, domain=-3:-1]  {1/(1+exp(-x))};
    \addplot[black!30, thin, domain=-1:-0.5]   {1/(1+exp(-x))};
    % Upper bound: chord through endpoints (convex => chord lies above)
    \addplot[blue, thick, dashed, domain=-3:-1] {0.1108*x + 0.3798};
    % Lower bound: tangent at d=-2
    \addplot[red,  thick, dashed, domain=-3:-1] {0.1050*x + 0.3292};
    % Drop lines
    \addplot[gray, thin, dashed] coordinates {(-3,-0.13) (-3,0.0474)};
    \addplot[gray, thin, dashed] coordinates {(-1,-0.13) (-1,0.2689)};
    \node[below] at (axis cs:-3,-0.13) {$l$};
    \node[below] at (axis cs:-1,-0.13) {$u$};
    \end{axis}
    \end{tikzpicture}
    \caption{$u\!\le\!0$: $h_U$ chord, $h_L$ tangent}
  \end{subfigure}
  \hfill
  % ── Case 3: unstable (l < 0 < u, mixed region) ───────────────────────────
  \begin{subfigure}[b]{0.32\textwidth}
    \centering
    \begin{tikzpicture}
    \begin{axis}[
        width=\textwidth,
        yticklabel=\empty, xticklabel=\empty,
        axis lines=middle, axis line style={latex-latex},
        axis on top=true,
        xmin=-3.5, ymin=-0.2, xmax=4.5, ymax=1.35,
        samples=100, clip=false,
        xlabel={\(x\)},
    ]
    % Shaded region between h_L and h_U
    \addplot[fill=blue!15, draw=none, forget plot]
        coordinates {(-2,0.0966) (3,0.9526) (3,1.1462) (-2,0.1192)};
    % Sigmoid
    \addplot[black, very thick, domain=-2:3] {1/(1+exp(-x))};
    % Upper bound: passes through (l,sigma(l)), tangent at d≈0.9
    \addplot[blue, thick, dashed, domain=-2:3] {0.2054*x + 0.5300};
    % Lower bound: passes through (u,sigma(u)), tangent at d≈-1.27
    \addplot[red,  thick, dashed, domain=-2:3] {0.1712*x + 0.4390};
    % Drop lines
    \addplot[gray, thin, dashed] coordinates {(-2,-0.13) (-2,0.1192)};
    \addplot[gray, thin, dashed] coordinates {(3,-0.13) (3,0.9526)};
    \node[below] at (axis cs:-2,-0.13) {$l$};
    \node[below] at (axis cs: 3,-0.13) {$u$};
    \end{axis}
    \end{tikzpicture}
    \caption{$\ell\!<\!0\!<\!u$: both bounds tangent}
  \end{subfigure}
  \caption{Linear relaxation bounds for $\sigma(x) = 1/(1+e^{-x})$ over
    $[\ell,u]$ (blue dashed: $h_U$; red dashed: $h_L$). Case~1 ($0 \le \ell$):
    $\sigma$ is concave, so $h_U$ is a tangent line and $h_L$ is the chord
    through the endpoints. Case~2 ($u \le 0$): $\sigma$ is convex and the
    roles are swapped. Case~3 ($\ell < 0 < u$): $\sigma$ transitions from
    convex to concave; $h_U$ passes through $(\ell,\sigma(\ell))$ and is tangent
    at some $d \ge 0$, while $h_L$ passes through $(u,\sigma(u))$ and is
    tangent at some $d \le 0$, found via binary search.}
  \label{fig:sigmoid}
\end{figure}

For a sigmoid neuron $\sigma(z) = \frac{e^z}{e^z + 1}$ with pre-activation bounds $\ell \le z \le u$, we seek linear functions $\hat{\sigma}_L(z)$ and $\hat{\sigma}_U(z)$ satisfying $\hat{\sigma}_L(z) \le \sigma(z) \le \hat{\sigma}_U(z)$ for all $z \in [\ell, u]$. Denote the left endpoint as $(\ell, \sigma(\ell))$ and the right endpoint as $(u, \sigma(u))$.
 
As with ReLU, there are three cases determined by the signs of $l$ and $u$. When $0 \le \ell \le u$, the sigmoid is concave over $[\ell, u]$, meaning the chord connecting both endpoints lies below the curve. The tightest lower bound $\hat{\sigma}_L$ is therefore the chord through the two endpoints, while the tightest upper bound $\hat{\sigma}_U$ is a tangent line to $\sigma$ at some point $d_U \in [\ell, u]$.

When $\ell \le u \le 0$, the sigmoid is convex over $[\ell, u]$, so the chord lies above the curve. Hence, the chord gives the tightest upper bound $\hat{\sigma}_U$, while the tightest lower bound $\hat{\sigma}_L$ is a tangent line at some $d_L \in [\ell, u]$.

When $\ell < 0 < u$, the neuron is unstable and $\sigma$ transitions from convex to concave over $[\ell, u]$. Here, the chord through the two endpoints crosses the curve and cannot serve as a sound bound. Instead, the upper bound $\hat{\sigma}_U$ is taken as the tangent line to $\sigma$ that also passes through the left endpoint $(\ell, \sigma(\ell))$, with tangent point $(d_U, \sigma(d_U))$ for some $d_U \ge 0$. Similarly, the lower bound $\hat{\sigma}_L$ is the tangent line passing through the right endpoint $(u, \sigma(u))$, with tangent point $(d_L, \sigma(d_L))$ for some $d_L \le 0$.

Due to the exponential form of $\sigma$, the tangent point $d$ cannot be solved for algebraically in any of the three cases. In practice, $d$ is found via binary search over a precomputed lookup table of $\sigma$ values.

\subsection{Bound Concretization}
\label{sec:concretization}

Once the linear coefficients have been computed, they are concretized into numerical bounds by optimizing over the input box $\mathcal{X}$:
\begin{equation}
    h_L = \min_{x \in \mathcal{X}}\; W_L x + b_L, \qquad h_U = \max_{x \in \mathcal{X}}\; W_U x + b_U.
    \label{eq:concreteOpt}
\end{equation}
Since $W_L x + b_L$ and $W_U x + b_U$ are linear in $x$ and $\mathcal{X}$  is a box constraint $[\ell, u]$, these optimizations yield closed-form solutions. Each linear objective over a box is minimized (or maximized) by selecting, for each input dimension, the endpoint that yields the smallest (or largest) contributions. By separating the weight matrices into their positive and negative components we get
\begin{equation}
    \min_{x \in \mathcal{X}} W_L x + b_L = W_L^+ \ell + W_L^- u + b_L, \qquad \max_{x \in \mathcal{X}} W_U x + b_U = W_U^+ u + W_U^- \ell + b_U,
    \label{eq:closedFormConcrete}
\end{equation}
where $W^+ = \max(W, 0)$ and $W^- = \min(W, 0)$ are applied element-wise.

\subsection{Optimized Bound Propagation via \alphaCROWN}
\label{sec:alpha-crown}

The \CROWN analysis relies on fixed heuristics to choose the relaxation slopes for unstable neurons. For ReLU (Section~\ref{app:relu_relaxation}), the lower bound slope $\alpha$ is set to either~$0$ or~$1$ depending on whether $u \geq |\ell|$. For the sigmoid relaxations (Section~\ref{app:sigmoid_relaxation}), the tangent points defining the bounds are determined by a fixed procedure. While computationally cheap, these fixed choices directly affect the tightness of the final concrete bounds.

Poor choices of relaxation slopes at these layers result in looser approximations that compound through subsequent layers, yielding increasingly loose bounds at the output. Although the fixed heuristics are computationally cheap, it can be advantageous to dynamically adjust the values of $\alpha$. Since verification asks whether the output bounds certify a property, even small changes in bound tightness can determine whether verification succeeds.

\alphaCROWN~\citep{xu2020automatic} generalizes \CROWN by observing that many nonlinear functions have more than one valid linear relaxation. Rather than choosing a fixed relaxation via a heuristic, \alphaCROWN parametrizes the choice of relaxation slope by a continuous variable $\alpha$, and optimizes it to find the tightest possible relaxation. 

\subsubsection{Parameterized Relaxations}
\label{sec:parametrized-relaxations}
\begin{figure}[t]
  \centering
  \begin{tikzpicture}
  \begin{axis}[
      width=0.48\textwidth,
      height=0.38\textwidth,
      yticklabel=\empty,
      xticklabel=\empty,
      axis lines=middle,
      axis line style={latex-latex},
      xmin=-4.5, ymin=-4, xmax=5.5, ymax=5,
      samples=100,
      clip=false,
      xlabel={\(x\)},
      every axis x label/.style={at={(ticklabel* cs:1)}, anchor=west},
  ]
  \addplot[black, very thick, samples at={-3, 0, 4}] {max(0,x)};
  \addplot[blue, very thick, dashed, domain=-3:4] {(4/7)*(x + 3)};
  \addplot[red, thick, loosely dashed, domain=-3:4] {0};
  \node[red, right, font=\footnotesize] at (axis cs:4.1, -0.45) {$\alpha = 0$};
  \addplot[red, thick, densely dashed, domain=-3:4] {0.25*x};
  \node[red, right, font=\footnotesize] at (axis cs:4.1, 1.0) {$\alpha = 0.25$};
  \addplot[red, thick, dash dot, domain=-3:4] {0.5*x};
  \node[red, right, font=\footnotesize] at (axis cs:4.1, 2.0) {$\alpha = 0.5$};
  \addplot[red, thick, densely dotted, domain=-3:4] {0.75*x};
  \node[red, right, font=\footnotesize] at (axis cs:4.1, 3.0) {$\alpha = 0.75$};
  \addplot[red, very thick, solid, domain=-3:4] {x};
  \node[red, right, font=\footnotesize] at (axis cs:4.1, 4.0) {$\alpha = 1$};
  \addplot[gray, thin, dashed] coordinates {(-3,-3.4) (-3,0)};
  \addplot[gray, thin, dashed] coordinates {(4,-3.4) (4,4)};
  \node[below, font=\small] at (axis cs:-3,-3.4) {$l$};
  \node[below, font=\small] at (axis cs: 4,-3.4) {$u$};
  \end{axis}
  \end{tikzpicture}
  \caption{Family of valid linear relaxations for an unstable ReLU neuron with pre-activation bounds $[l,u]$. The upper bound (blue dashed) is uniquely determined as the line connecting the points $(l,0)$ and $(u,u)$. The lower bound (red) can be given by any valid slope, $\alpha \in [0,1]$. Standard \CROWN selects $\alpha$ via heuristic choice, while \alphaCROWN  treats each $\alpha_i$ as an optimizable parameter.}
  \label{fig:relu_relaxation}
\end{figure}

Consider an unstable ReLU neuron with pre-activation bounds $\ell < 0 < u$. Recall from \eqref{eq:reluRelax} that the upper bound is uniquely determined as the line connecting $(\ell, 0)$ and $(u, u)$; however, the lower bound is not unique. Any line of the form of $h_L(z) = \alpha z$ with $\alpha \in [0, 1]$ satisfies $h_L(z) \leq \max(0, z)$ for all $z \in [\ell, u]$. Geometrically, $\alpha$ controls the slope of the lower bound line through the origin. At $\alpha = 0$, the lower bound is the horizontal line $h_L=0$; at $\alpha = 1$ it is the identity $h_L=z$. Intermediate values of $\alpha$ control the tilt of the lower bound between these two extremes, as seen in Figure~\ref{fig:relu_relaxation}. 

\alphaCROWN assigns a separate parameter $\alpha_i$ to each unstable neuron $i$. By collecting all the $\alpha$'s into a single vector $\alpha = (\alpha_1, \ldots, \alpha_k)$ where $k$ is the number of unstable neurons, the back substitution becomes a function of both the input $x$ and $\alpha$. The concretized output bounds from~\eqref{eq:concreteOpt} therefore become
\begin{equation}
    L(x, \alpha) \;\leq\; h^{\text{output}}(x) \;\leq\; U(x, \alpha),
    \label{eq:alpha_bounds}
\end{equation}
where $L(x, \alpha)$ and $U(x, \alpha)$ are linear in $x$ for any fixed $\alpha$.
 
A similar parametrization extends to sigmoid and other activation functions. However, unlike ReLU neurons where only the lower bounds admits a family of valid relaxation and the upper bound is fixed, these activation often have freedom in both the upper and lower bounds. This typically requires two parameters per unstable neuron, $\alpha_i^L$ and $\alpha_i^U$, one for each bound. 

\subsubsection{Optimization via Projected Gradient Descent}
Given the parameterized bounds in \eqref{eq:alpha_bounds}, $\alpha$-CROWN seeks the relaxation parameters that yield the tightest concrete bounds. For the lower bound we want to maximize $L(x, \alpha)$ over $\alpha$. Conversely, for the upper bound we want to minimize $U(x, \alpha)$. This gives the optimization problems
\begin{equation}
    \max_{\alpha \in \mathcal{A}} \min_{x \in \mathcal{X}} L(x, \alpha), \qquad \min_{\alpha \in \mathcal{A}} \max_{x \in \mathcal{X}} U(x, \alpha),
    \label{eq:alpha_opt}
\end{equation}
where $\mathcal{A}$ denotes the set of valid relaxation parameters (e.g., $\alpha \in [0, 1]$ for each unstable ReLU neuron). The inner optimization over $x \in \mathcal{X}$ is solved in closed form via~\eqref{eq:closedFormConcrete}, so the problem reduces to an optimization over $\alpha$ alone. Note that the lower and upper bound optimizations are performed independently, each with their own $\alpha$ parameters, since the slope that produces the tightest lower, may differ from the lower slope that produces the tightest upper bound. 

\alphaCROWN solves~\eqref{eq:alpha_opt} using projected gradient descent (PGD)~\citep{madry2017towards}, starting from the initial \CROWN heuristic values. At each iteration, the $\alpha$ parameters are updated using the gradient of the concrete bounds with respect to $\alpha$, taking a step in the direction that tightens the bound, and projecting back onto the feasible set $\mathcal{A}$ (e.g., clamping each $\alpha_i$ to $[0, 1]$). After a fixed number of iterations, the optimized $\alpha^*$ are used in a final \CROWN backward pass to produce the tightened bounds. This iterative process produces tighter bounds than the \CROWN heuristics, at the cost of additional computation from the repeated passes. 

\section{Evaluation}

\subsection{Per-Instance CPU Results}
\label{app:per-instance-CPU-plots}
\begin{figure}[!ht]
    \centering
    \includegraphics[width=0.34\linewidth]{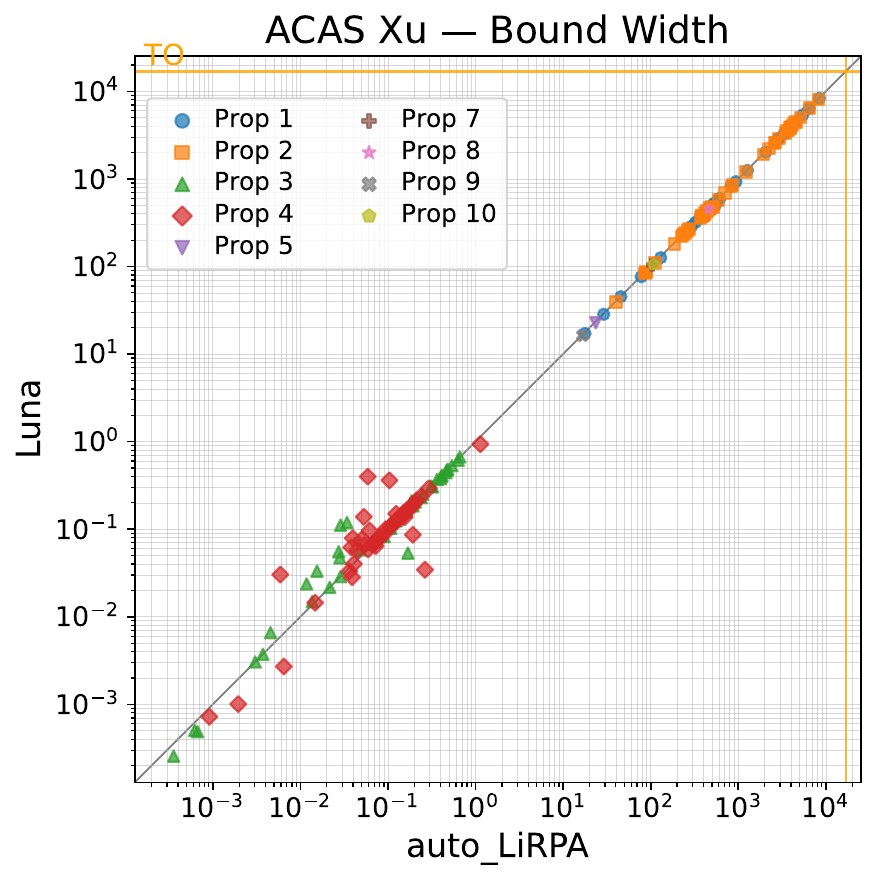}
    \hfil
    \includegraphics[width=0.34\linewidth]{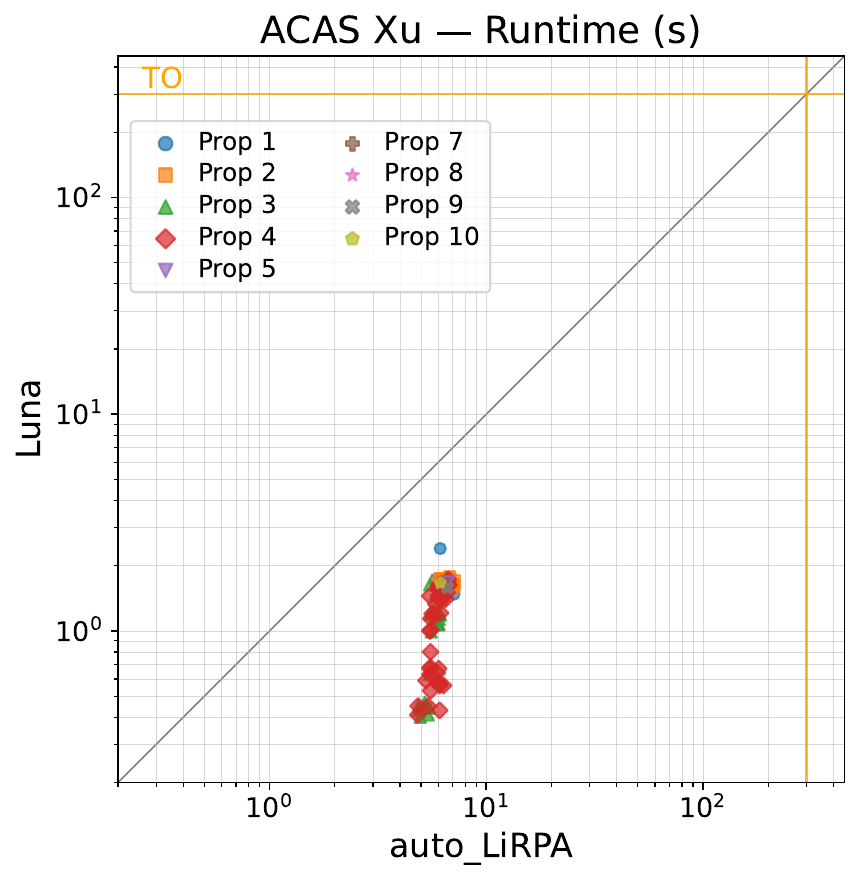}
    \caption{Bound width (left) and runtime (right) for Acasxu 2023}
    \label{fig:acasxu_2023}
\end{figure}

\begin{figure}[!ht]
    \centering
    \includegraphics[width=0.34\linewidth]{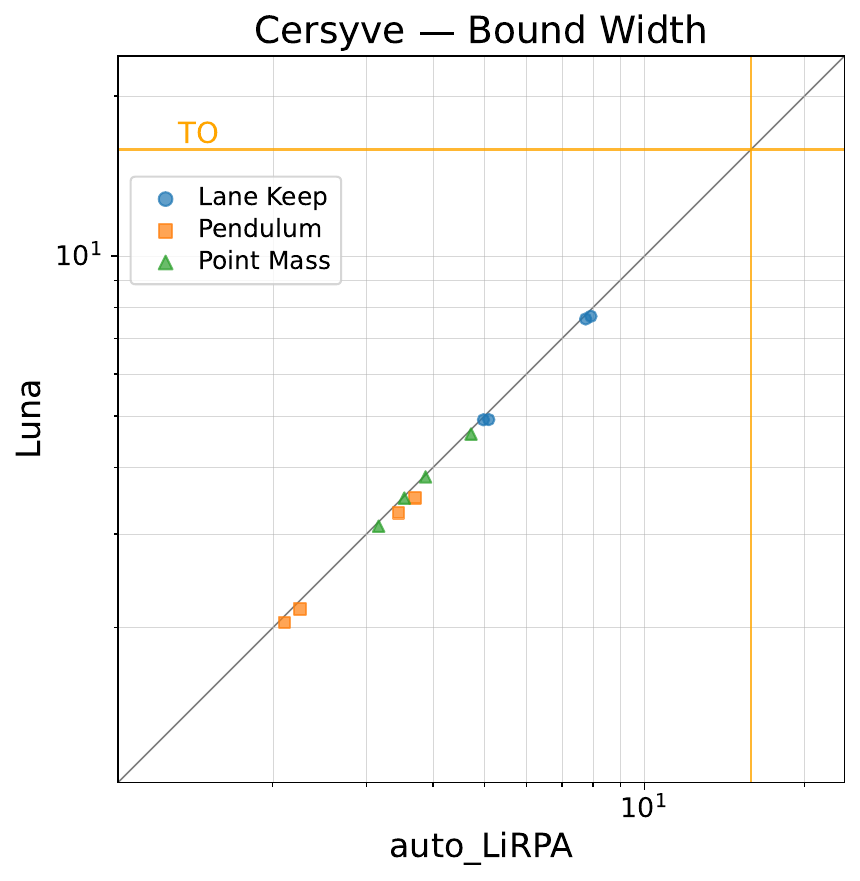}
    \hfil
    \includegraphics[width=0.34\linewidth]{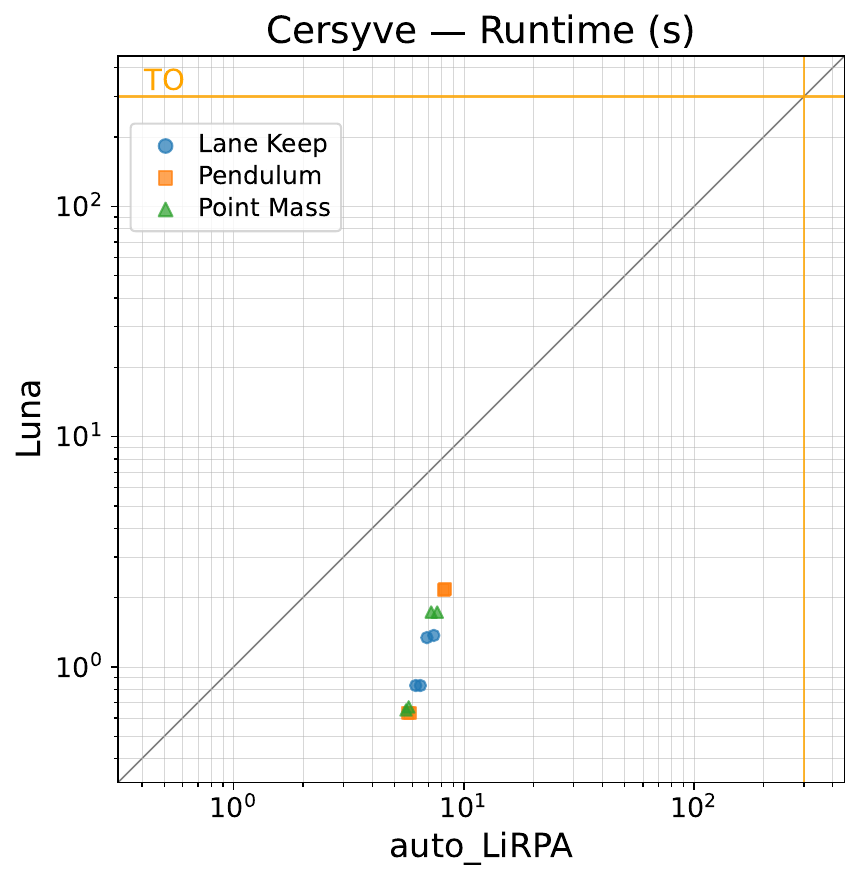}
    \caption{Bound width (left) and runtime (right) for Cersyve}
    \label{fig:cersyve}
\end{figure}

\begin{figure}[!ht]
    \centering
    \includegraphics[width=0.34\linewidth]{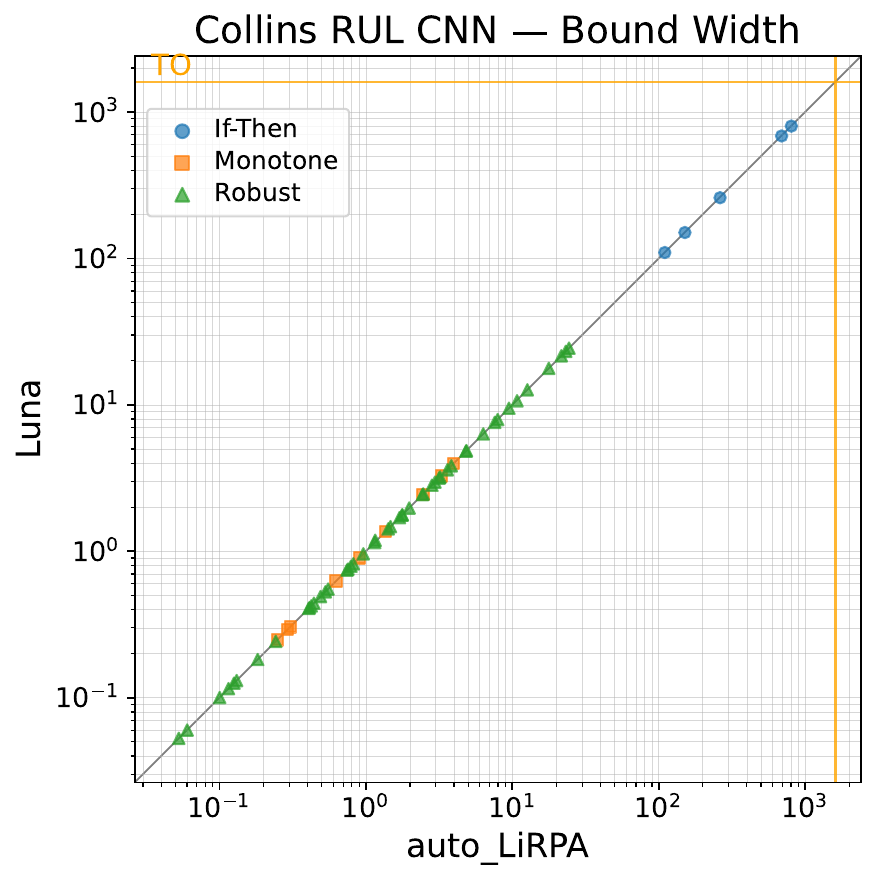}
    \hfil
    \includegraphics[width=0.34\linewidth]{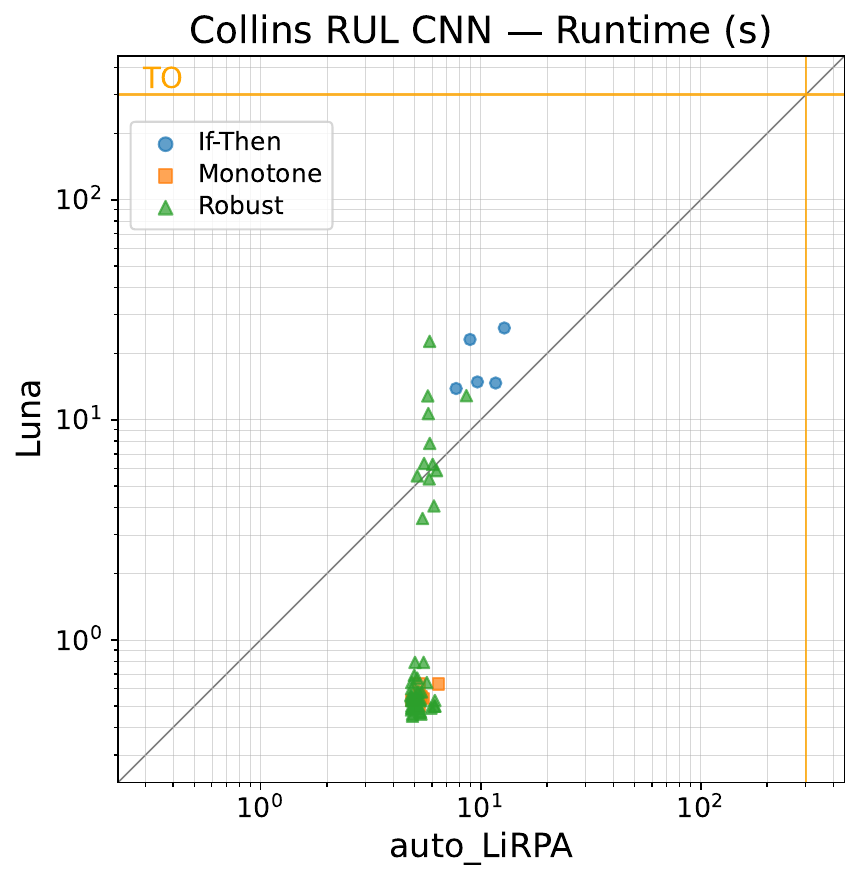}
    \caption{Bound width (left) and runtime (right) for Collins Rul Cnn 2022}
    \label{fig:collins_rul_cnn_2022}
\end{figure}

\begin{figure}[!ht]
    \centering
    \includegraphics[width=0.34\linewidth]{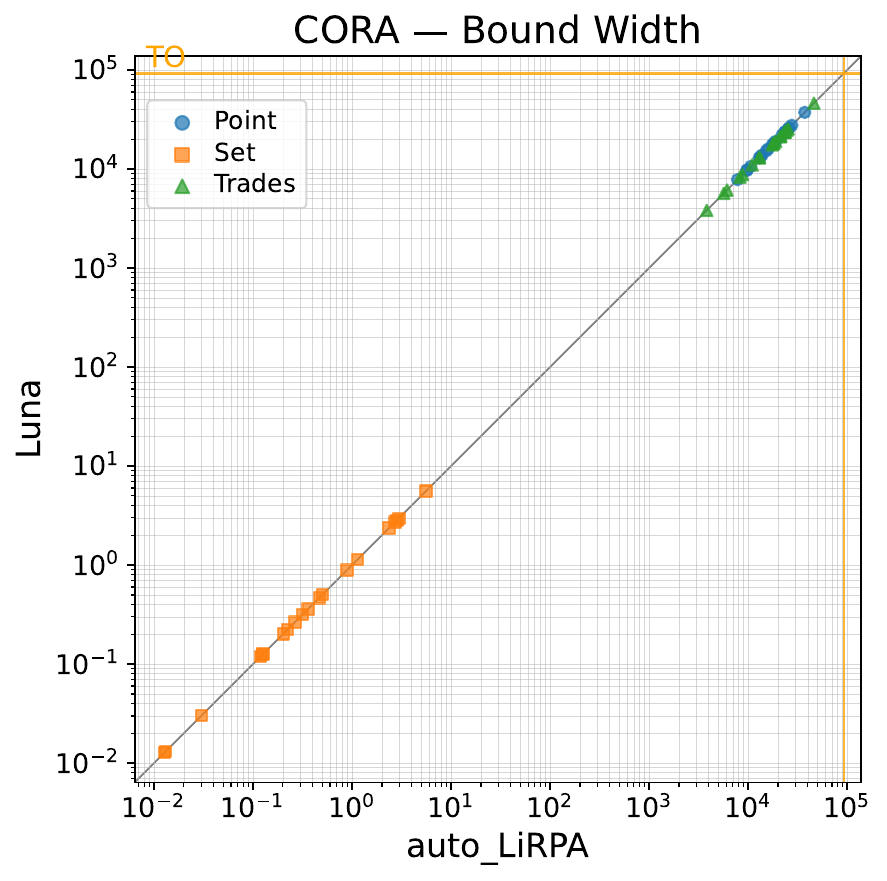}
    \hfil
    \includegraphics[width=0.34\linewidth]{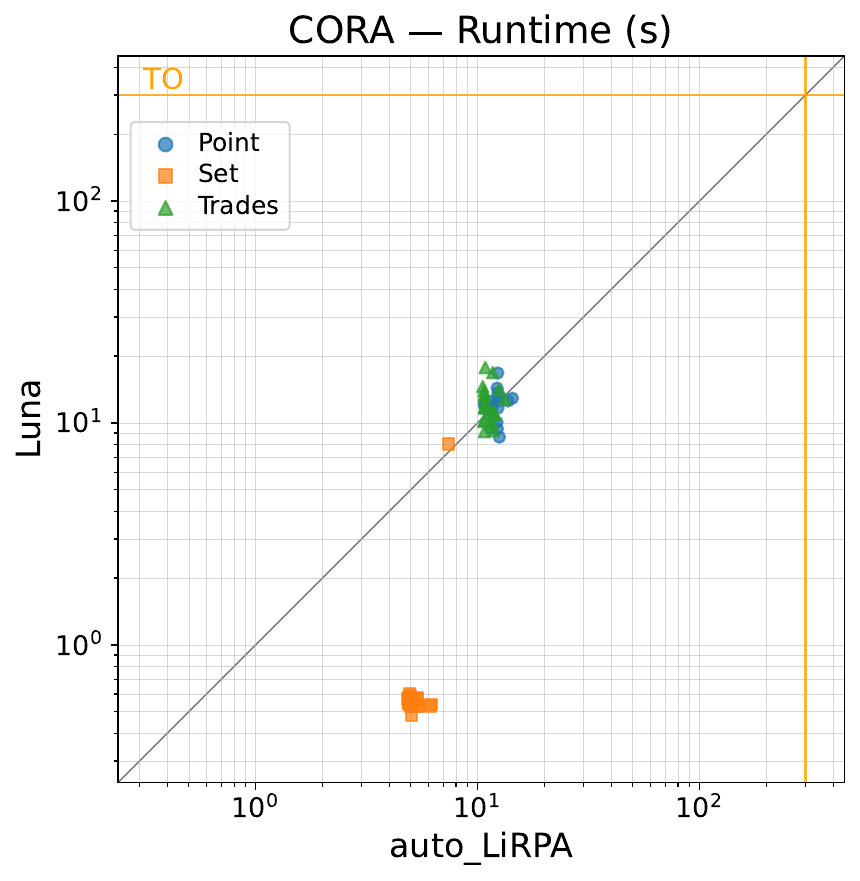}
    \caption{Bound width (left) and runtime (right) for Cora 2024}
    \label{fig:cora_2024}
\end{figure}

\begin{figure}[!ht]
    \centering
    \includegraphics[width=0.34\linewidth]{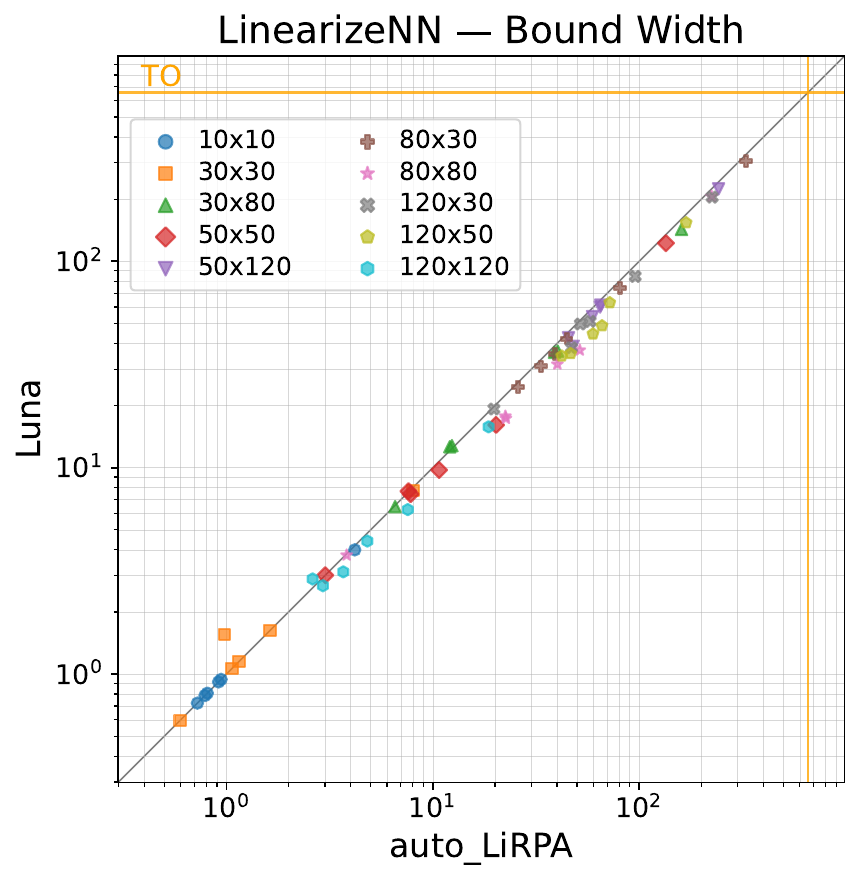}
    \hfil
    \includegraphics[width=0.34\linewidth]{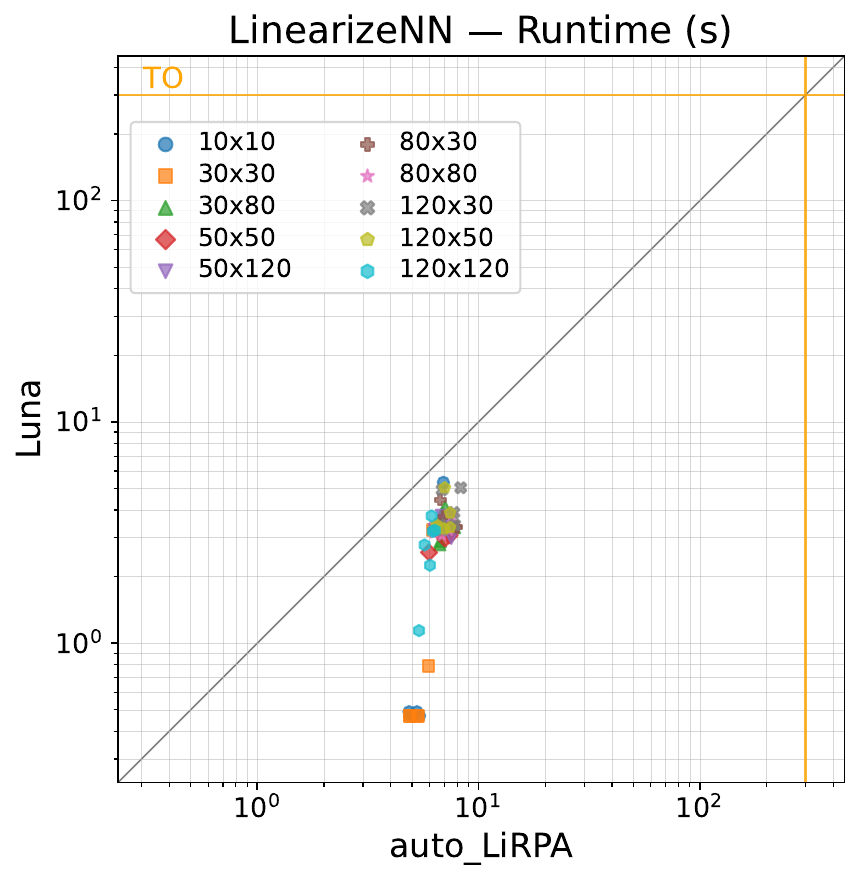}
    \caption{Bound width (left) and runtime (right) for Linearizenn 2024}
    \label{fig:linearizenn_2024}
\end{figure}

\begin{figure}[!ht]
    \centering
    \includegraphics[width=0.34\linewidth]{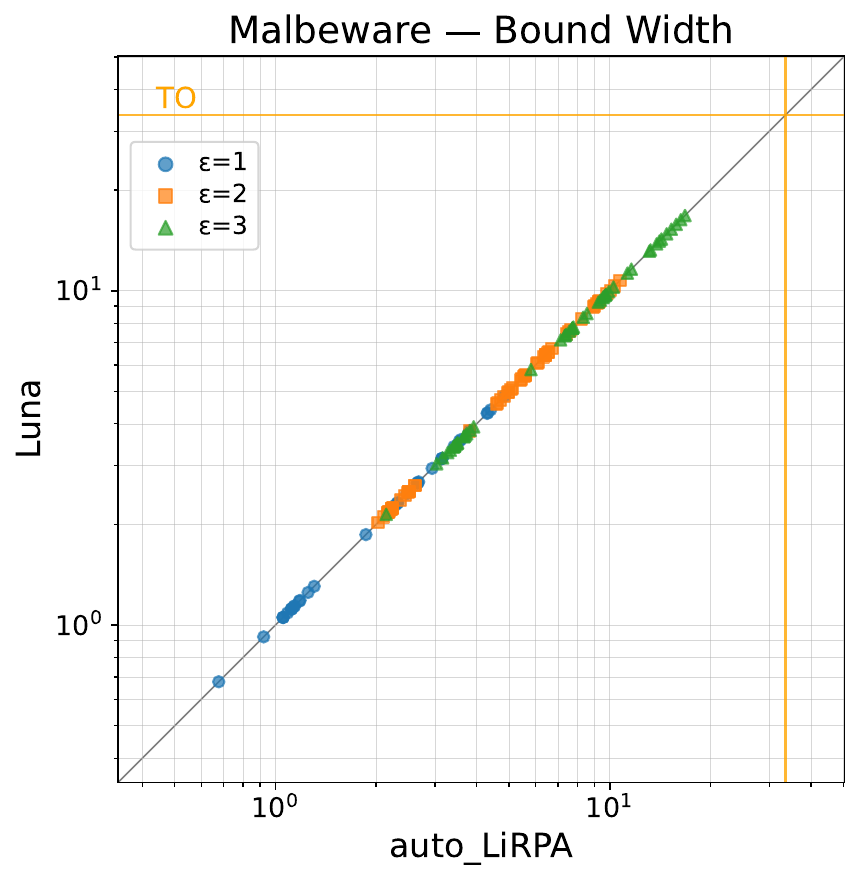}
    \hfil
    \includegraphics[width=0.34\linewidth]{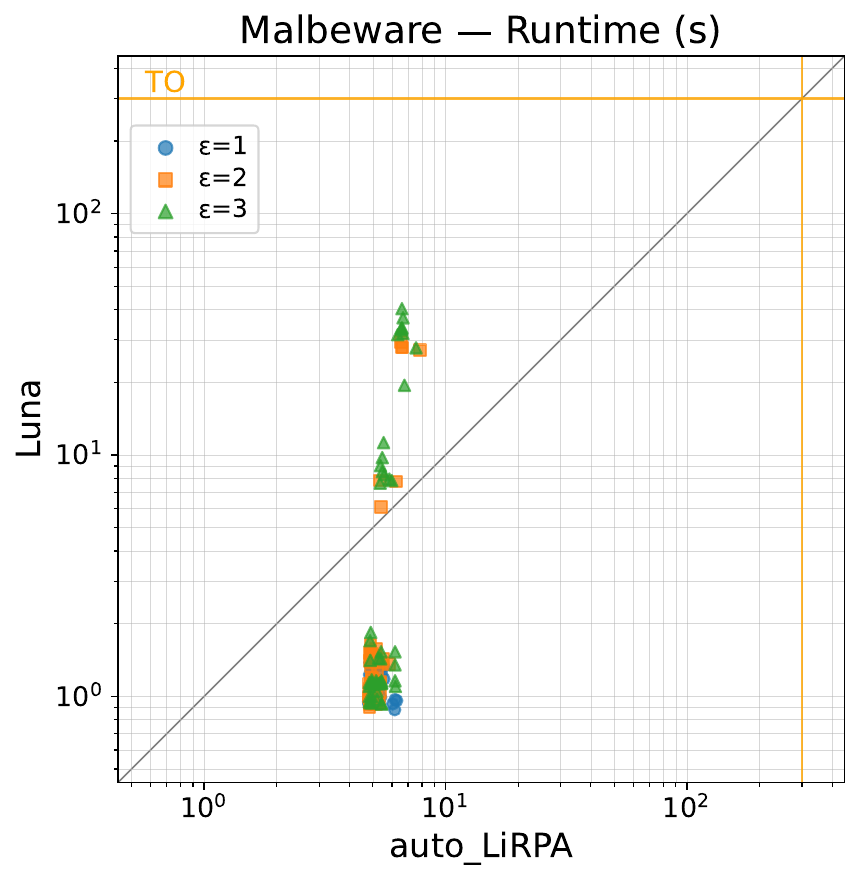}
    \caption{Bound width (left) and runtime (right) for Malbeware}
    \label{fig:malbeware}
\end{figure}

\begin{figure}[!ht]
    \centering
    \includegraphics[width=0.34\linewidth]{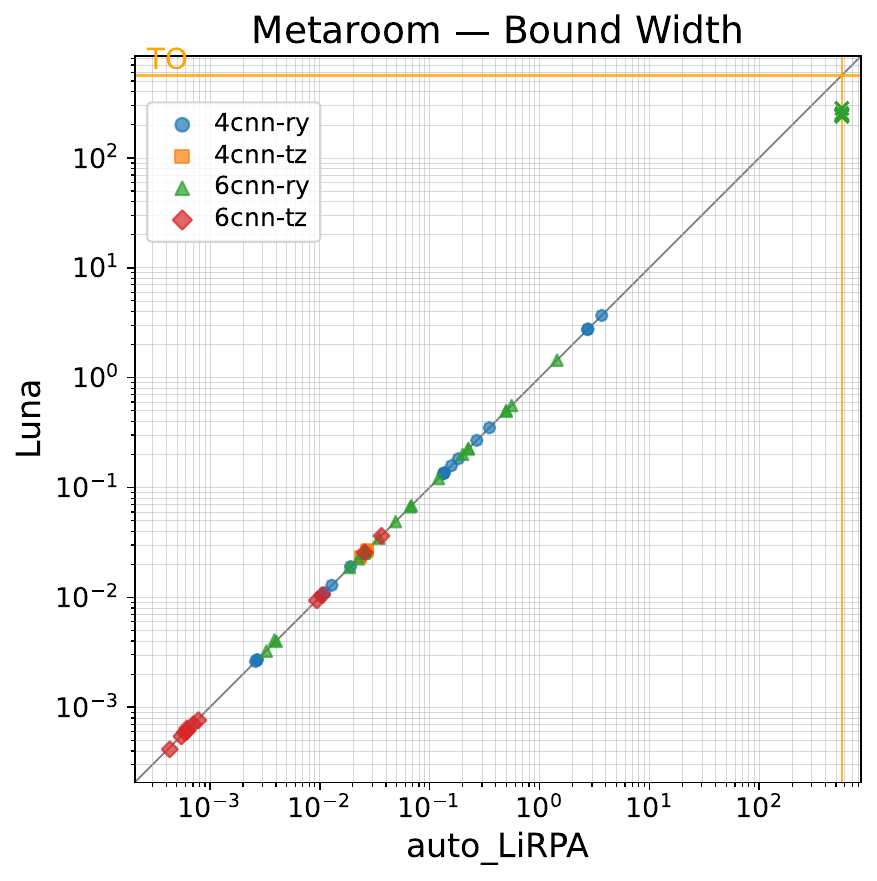}
    \hfil
    \includegraphics[width=0.34\linewidth]{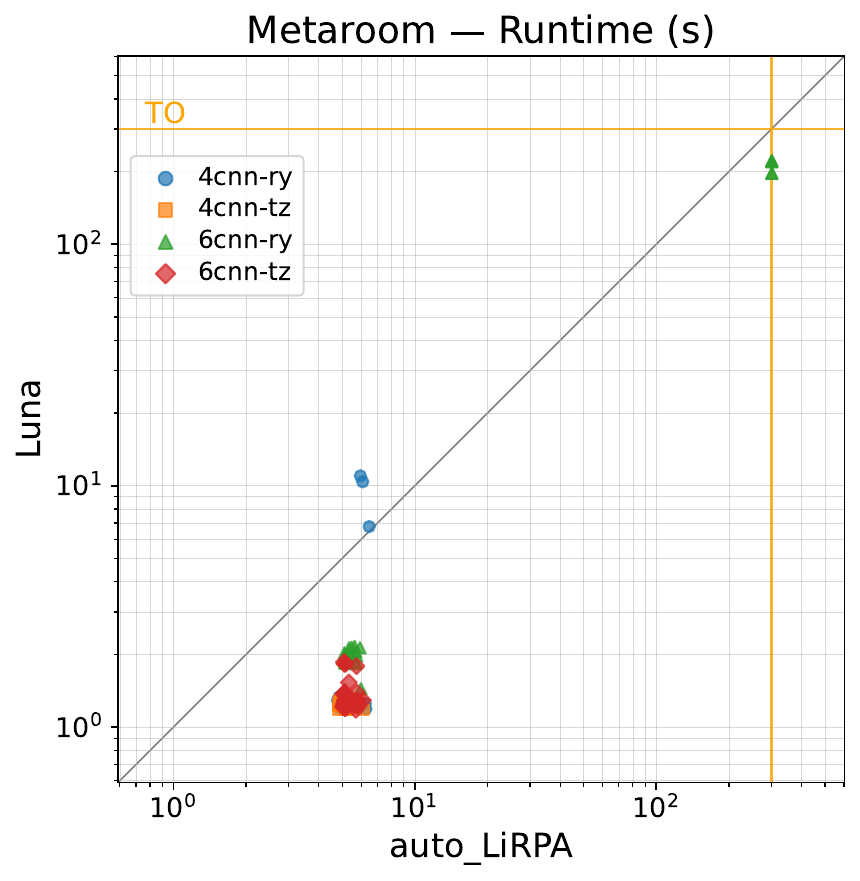}
    \caption{Bound width (left) and runtime (right) for Metaroom 2023}
    \label{fig:metaroom_2023}
\end{figure}

\begin{figure}[!ht]
    \centering
    \includegraphics[width=0.34\linewidth]{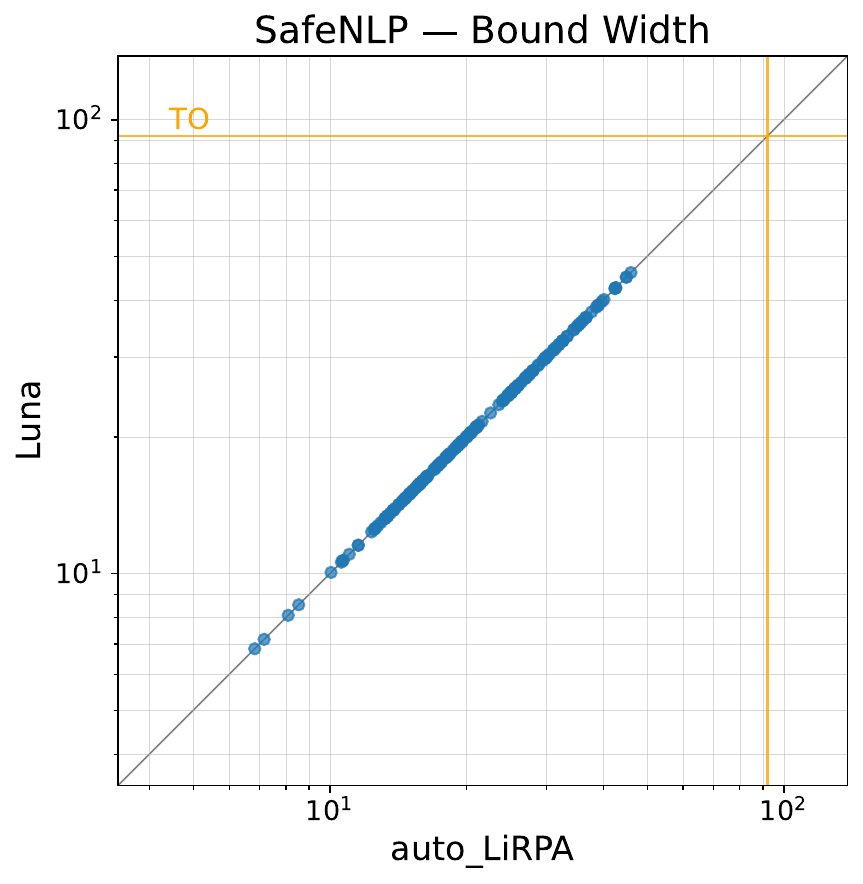}
    \hfil
    \includegraphics[width=0.34\linewidth]{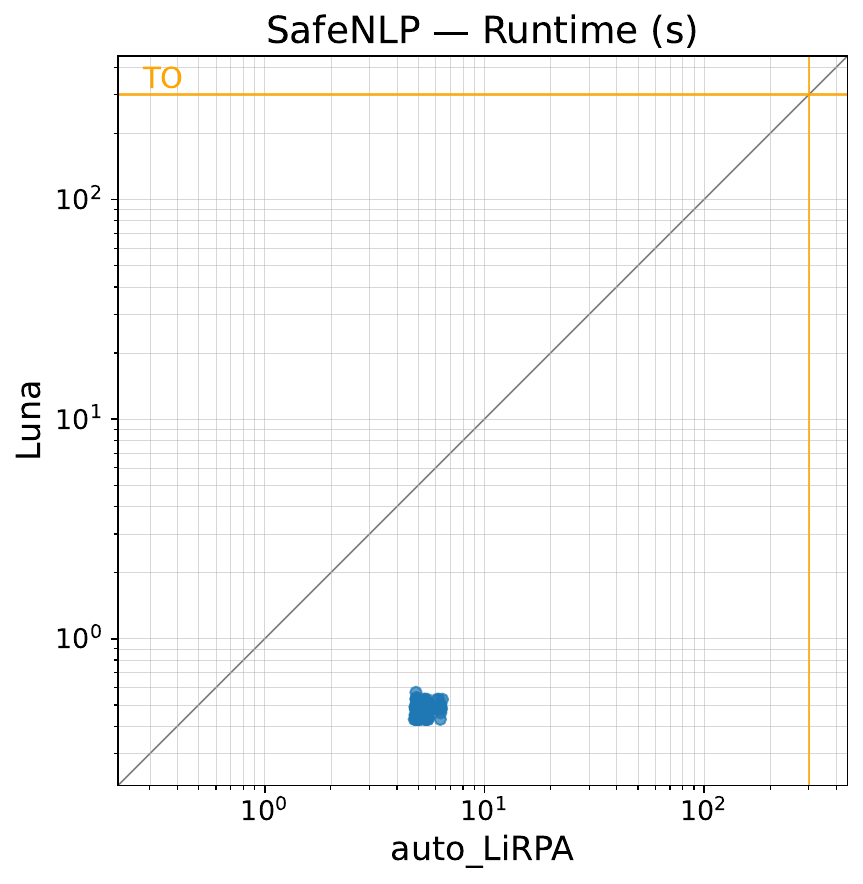}
    \caption{Bound width (left) and runtime (right) for Safenlp 2024}
    \label{fig:safenlp_2024}
\end{figure}

\begin{figure}[!ht]
    \centering
    \includegraphics[width=0.34\linewidth]{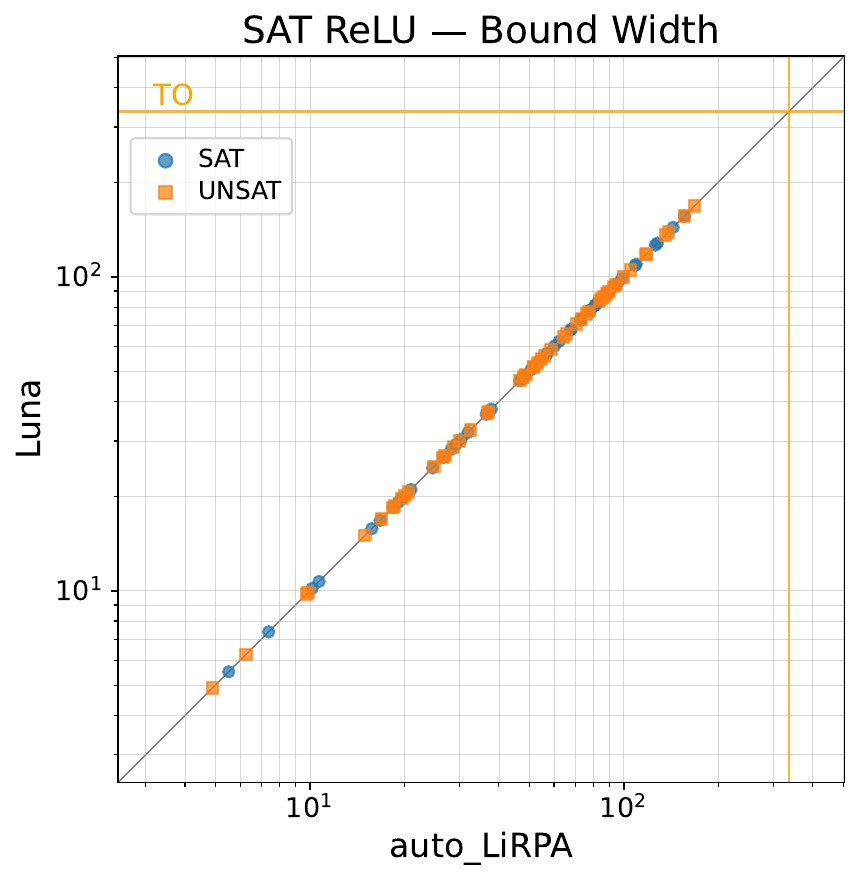}
    \hfil
    \includegraphics[width=0.34\linewidth]{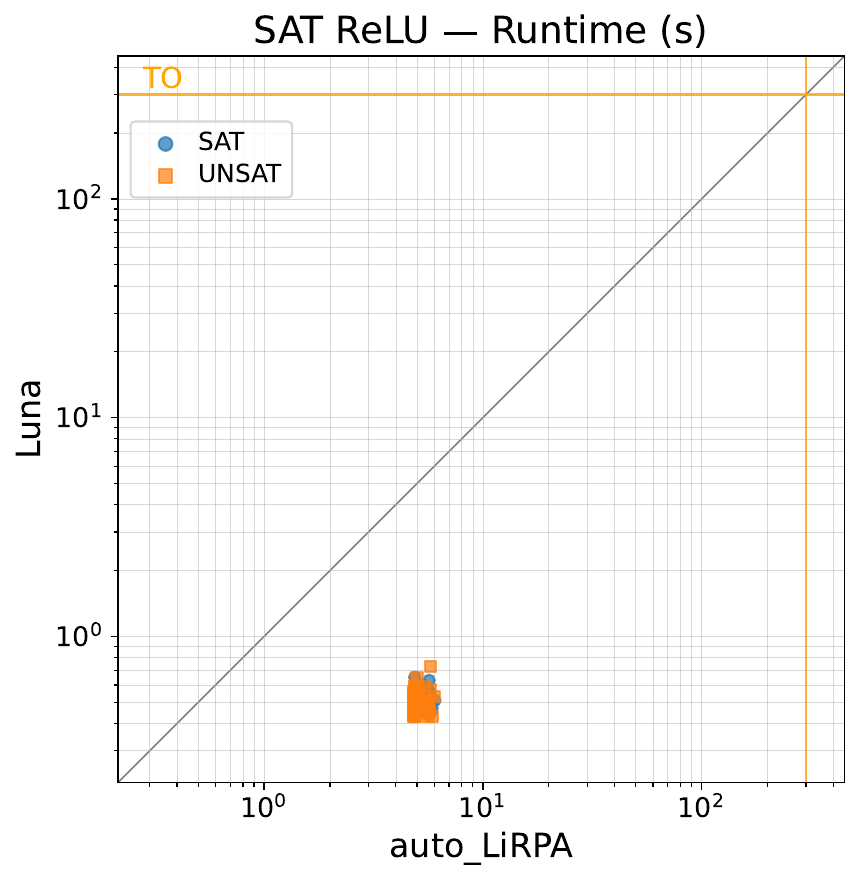}
    \caption{Bound width (left) and runtime (right) for Sat Relu}
    \label{fig:sat_relu}
\end{figure}

\begin{figure}[!ht]
    \centering
    \includegraphics[width=0.34\linewidth]{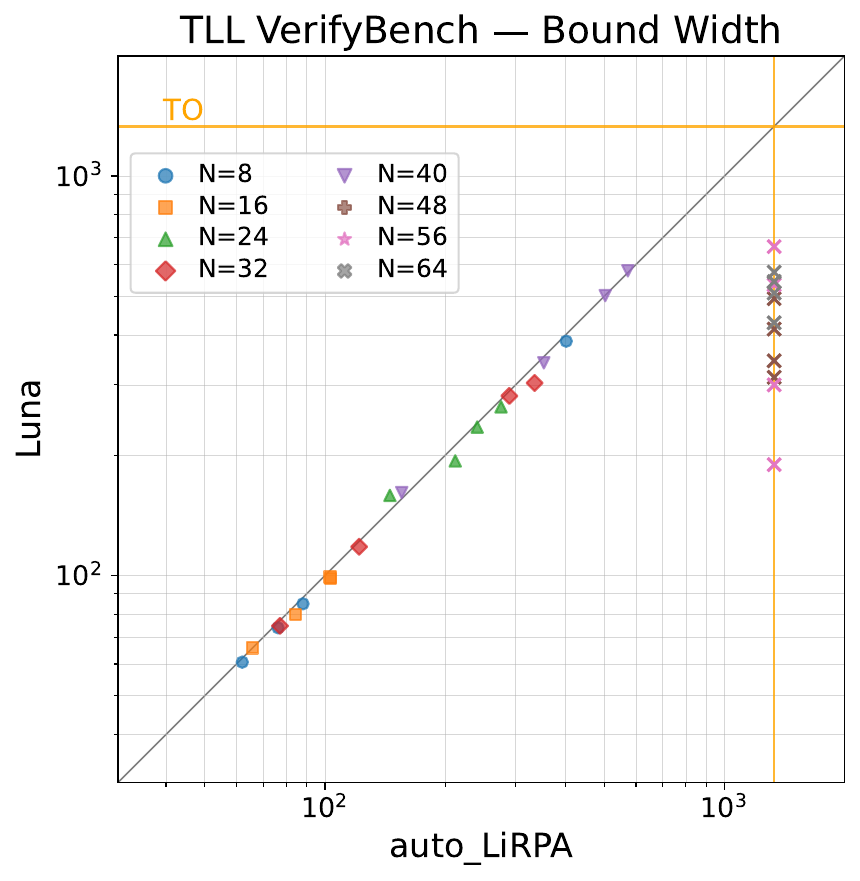}
    \hfil
    \includegraphics[width=0.34\linewidth]{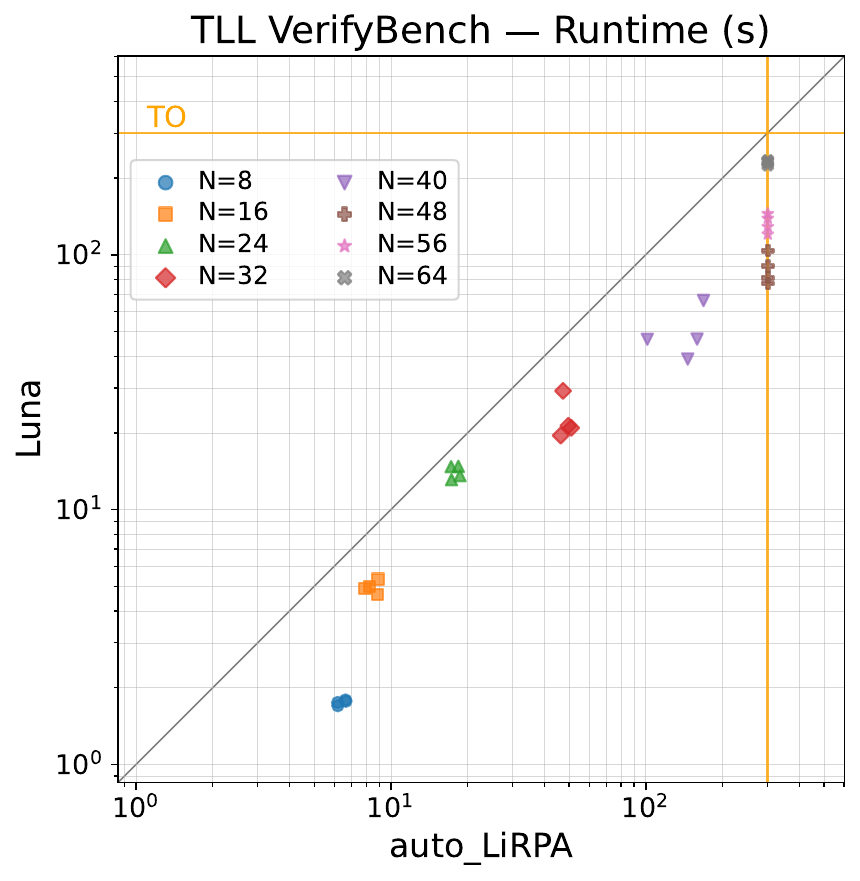}
    \caption{Bound width (left) and runtime (right) for Tllverifybench 2023}
    \label{fig:tllverifybench_2023}
\end{figure}

\Cref{fig:acasxu_2023,fig:cersyve,fig:collins_rul_cnn_2022,fig:cora_2024,fig:linearizenn_2024,fig:malbeware,fig:metaroom_2023,fig:safenlp_2024,fig:sat_relu,fig:tllverifybench_2023} show per-instance and per-property results for \sys and \autolirpa on the same set of evaluated benchmarks as Table~\ref{tab:bound_width_runtime}. In the bound-width plots, points below the diagonal line indicate that \sys produced tighter bounds, while points lying directly on the line indicate the tools computed identical average bounds for all output dimensions in that instance. Similarly, in the runtime plots, points below the line indicate that \sys was faster. A point on the timeout (TO) boundary indicates that the corresponding tool did not finish within the 300-second timeout. While the aggregate statistics in Table~\ref{tab:bound_width_runtime} summarize overall trends, the scatter plots show finer-grained patterns dependent on network size and property type that are not visible in the averages.

On several benchmarks the scatter plots reveal that bound-width and runtime differences are not uniform across instances but tend to be correlated with specific verification properties or sub-benchmark categories.
 
The acasxu\_2023 benchmark (\Cref{fig:acasxu_2023}, left), a suite of networks designed for aircraft collision avoidance, is the most notable example of property-dependent bound-width variation. On instances with bound widths above approximately $1.0$, both tools produce nearly identical results. Below this threshold, the spread noticeably increases. These lower-magnitude instances belong to Properties~3 and~4, which verify the network's behavior in close-range encounters where the correct advisory output must be distinguished from all other outputs by a small margin. The competing outputs are inherently close in value, producing smaller-magnitude bounds. When the bound being optimized is small, the \alphaCROWN loss surface is relatively flat near the optimum. The small-magnitude bounds and correspondingly small gradient signal make the optimization process more sensitive to implementation-level differences. These can include a different ordering of operations in the relaxation process, different values for intermediate bounds during optimization, or differences in how gradients are computed and accumulated. These differences in the optimization process cause \sys and \autolirpa to converge to different local optima.
 
A similar property-dependent pattern appears in runtime. On collins\_rul\_cnn\_2022 (\Cref{fig:collins_rul_cnn_2022}, right), the If-Then instances and a subset of the Robust instances show cases where \autolirpa achieves a faster runtime than \sys. \autolirpa is more efficient on all of the If-Then instances and on roughly half of the Robust instances. The different properties encode different specifications on the network's output, which affects the structure of the final verification problem. \sys handles output specifications by appending the specification matrix as an additional layer, treating all specifications uniformly. By contrast, \autolirpa has specialized constraint handling for common specification types such as one-hot encodings. This specialized handling likely reduces the cost of the backward pass for these property types. \sys targets the development of the vanilla \alphaCROWN analysis and does not yet implement these specification-specific optimizations.
 
On the remaining six benchmarks, cersyve (\Cref{fig:cersyve}), linearizenn\_2024 (\Cref{fig:linearizenn_2024}), metaroom\_2023 (\Cref{fig:metaroom_2023}), safenlp\_2024 (\Cref{fig:safenlp_2024}), sat\_relu (\Cref{fig:sat_relu}), and tllverifybench\_2023 (\Cref{fig:tllverifybench_2023}) the bound-width plots show that \sys is a competitive bound propagation engine with significant runtime speedups. For nearly all instances across these benchmarks, both tools compute similar or equivalent bounds, with no systematic patterns across properties or networks, while \sys achieves faster runtimes. 

\subsection{Per-Instance GPU Results}
\label{app:per-instance-GPU-plots}
\begin{figure}[!ht]
    \centering
    \includegraphics[width=0.33\linewidth]{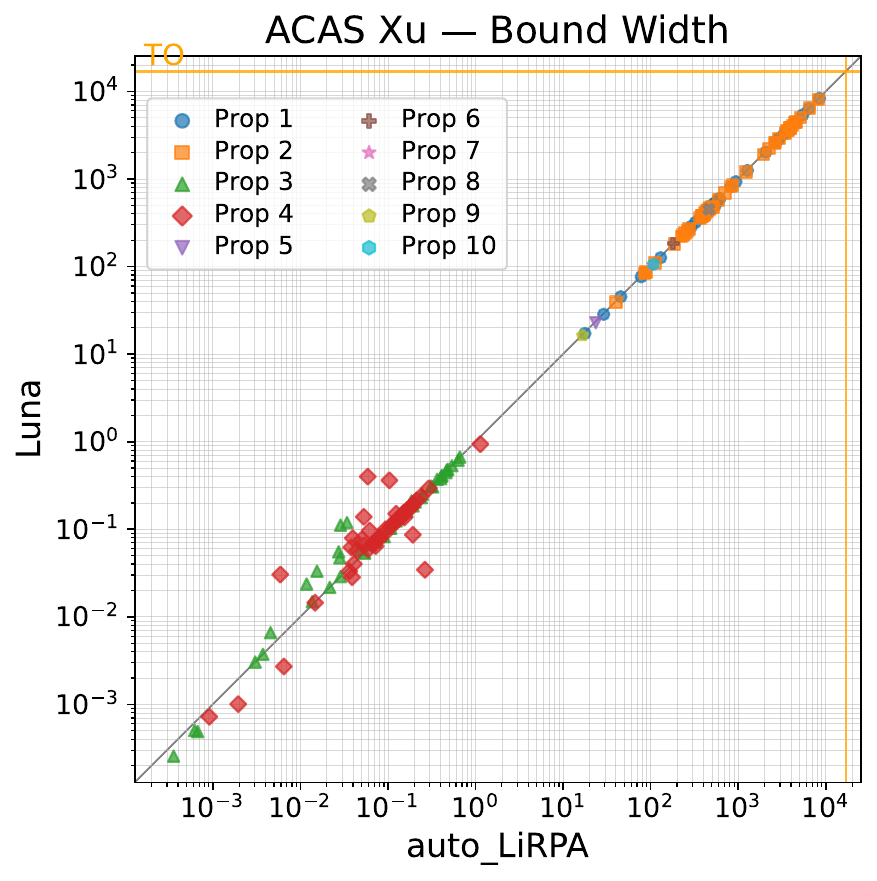}
    \hfil
    \includegraphics[width=0.33\linewidth]{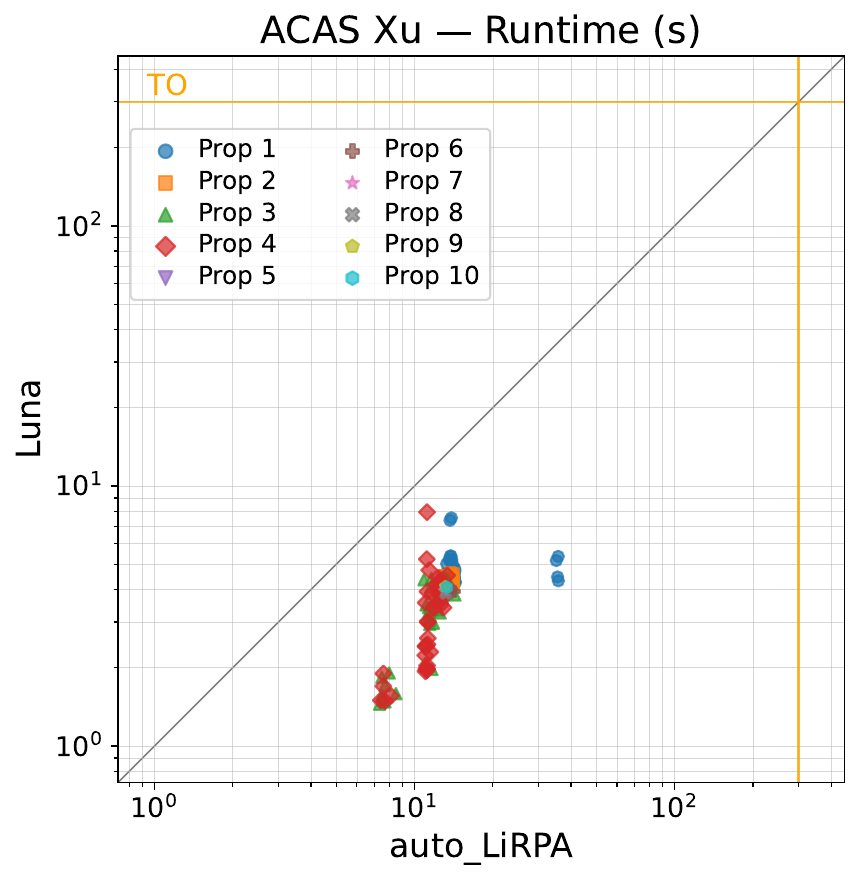}
    \caption{Bound width (left) and runtime (right) for Acasxu 2023 (GPU)}
    \label{fig:gpu_acasxu_2023}
\end{figure}

\begin{figure}[!ht]
    \centering
    \includegraphics[width=0.33\linewidth]{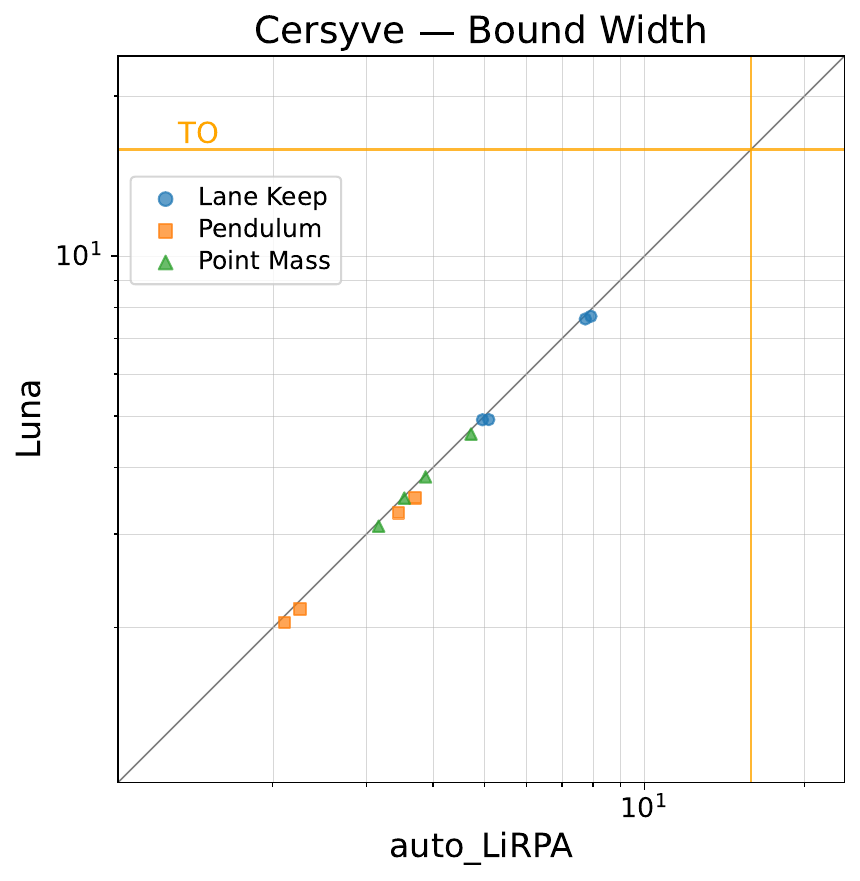}
    \hfil
    \includegraphics[width=0.33\linewidth]{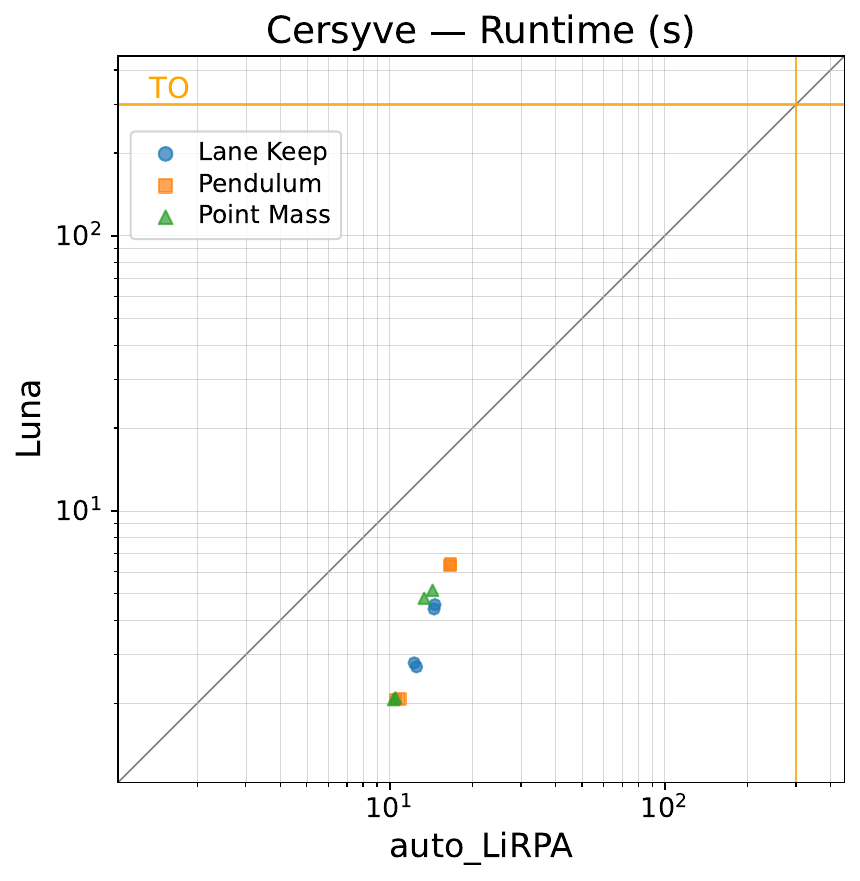}
    \caption{Bound width (left) and runtime (right) for Cersyve (GPU)}
    \label{fig:gpu_cersyve}
\end{figure}

\begin{figure}[!ht]
    \centering
    \includegraphics[width=0.33\linewidth]{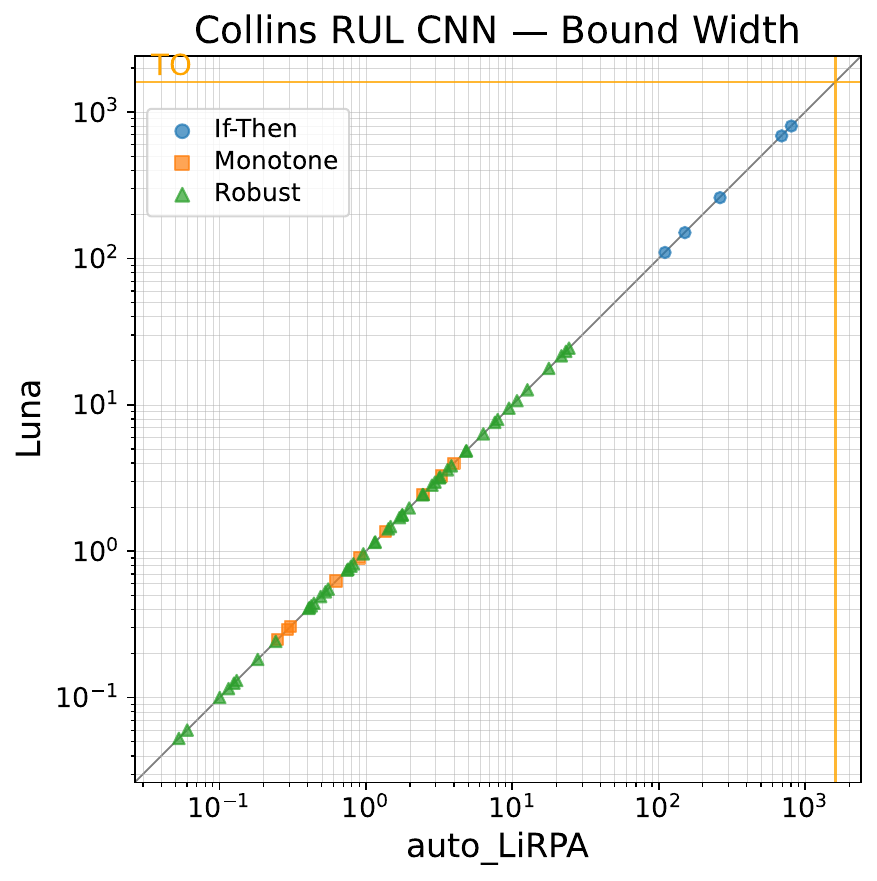}
    \hfil
    \includegraphics[width=0.33\linewidth]{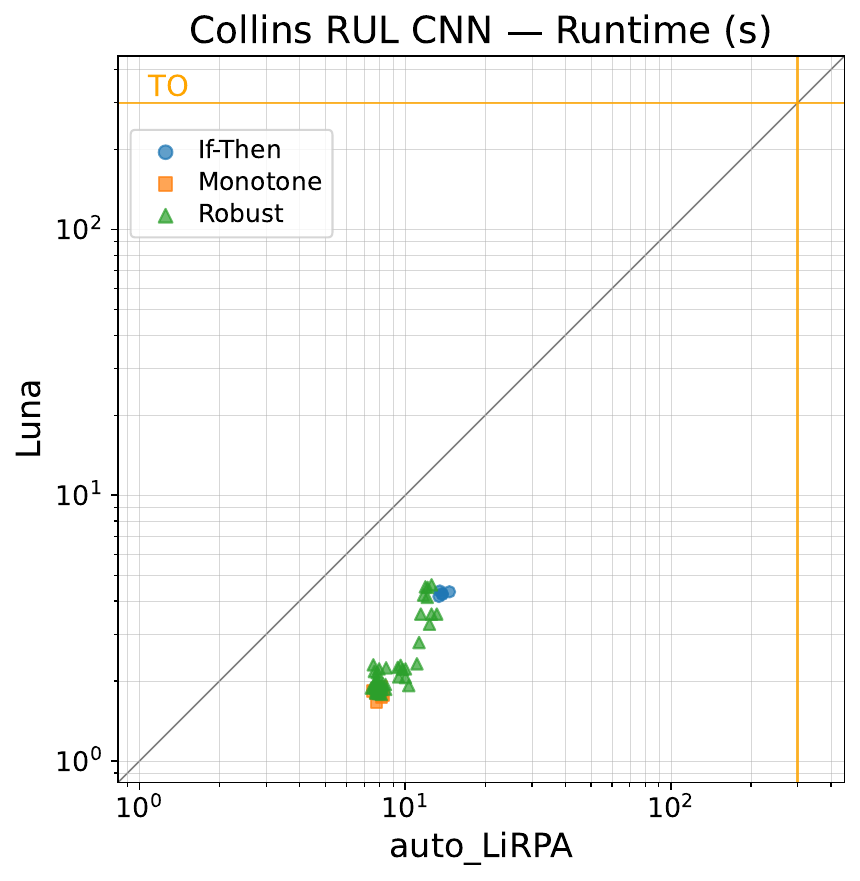}
    \caption{Bound width (left) and runtime (right) for Collins Rul Cnn 2022 (GPU)}
    \label{fig:gpu_collins_rul_cnn_2022}
\end{figure}

\begin{figure}[!ht]
    \centering
    \includegraphics[width=0.33\linewidth]{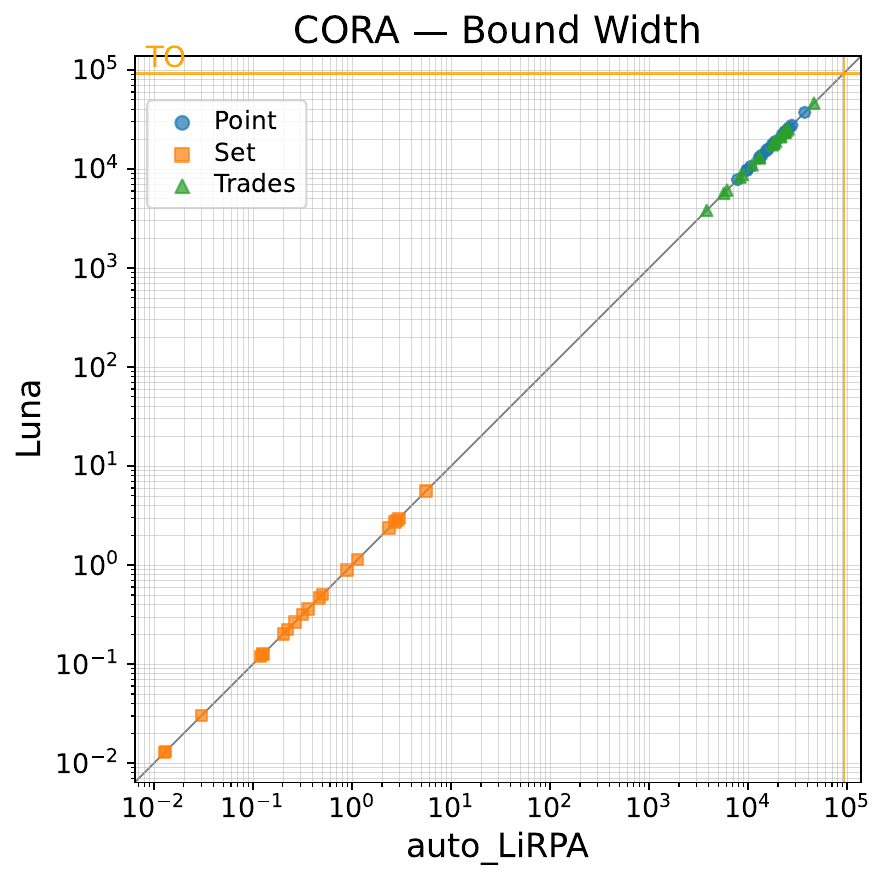}
    \hfil
    \includegraphics[width=0.33\linewidth]{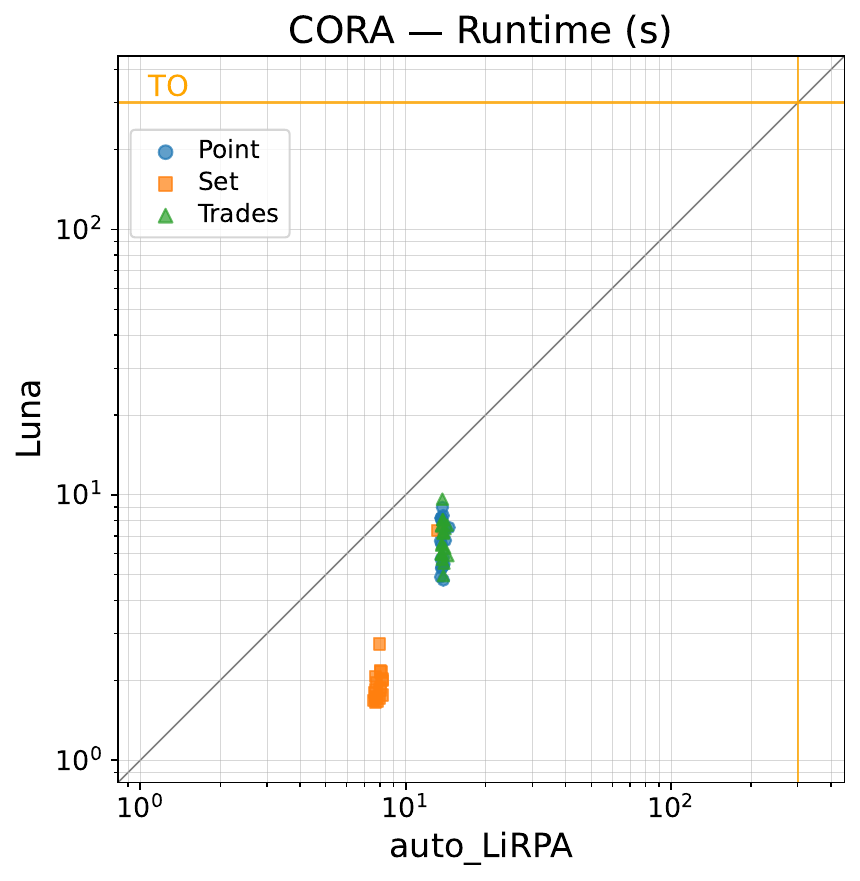}
    \caption{Bound width (left) and runtime (right) for Cora 2024 (GPU)}
    \label{fig:gpu_cora_2024}
\end{figure}

\begin{figure}[!ht]
    \centering
    \includegraphics[width=0.33\linewidth]{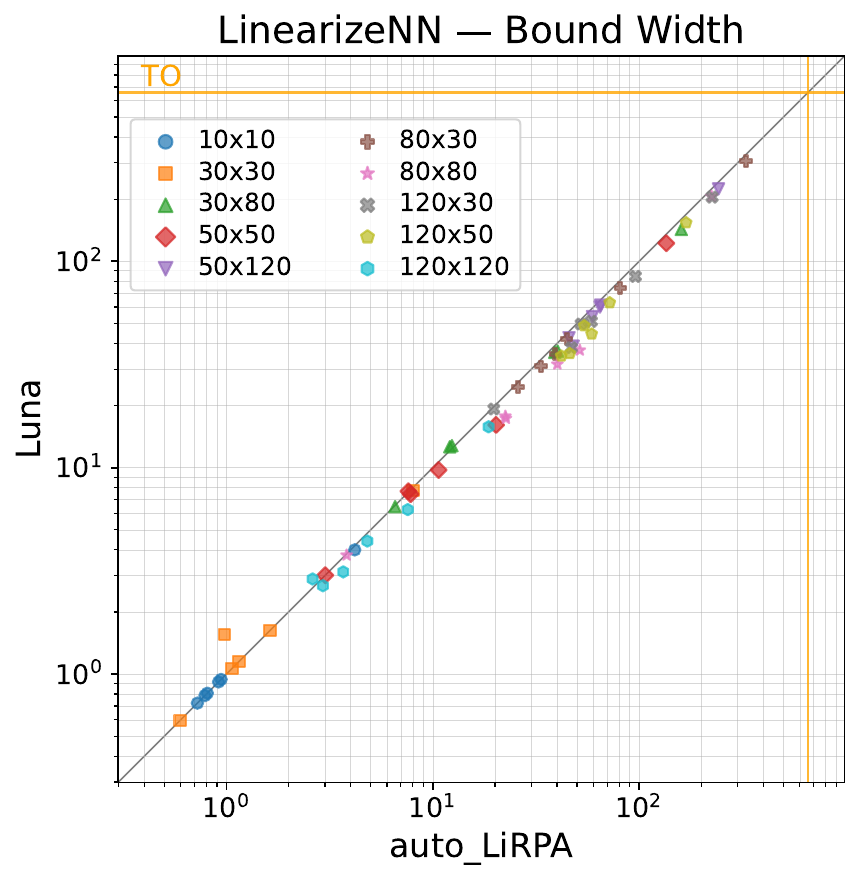}
    \hfil
    \includegraphics[width=0.33\linewidth]{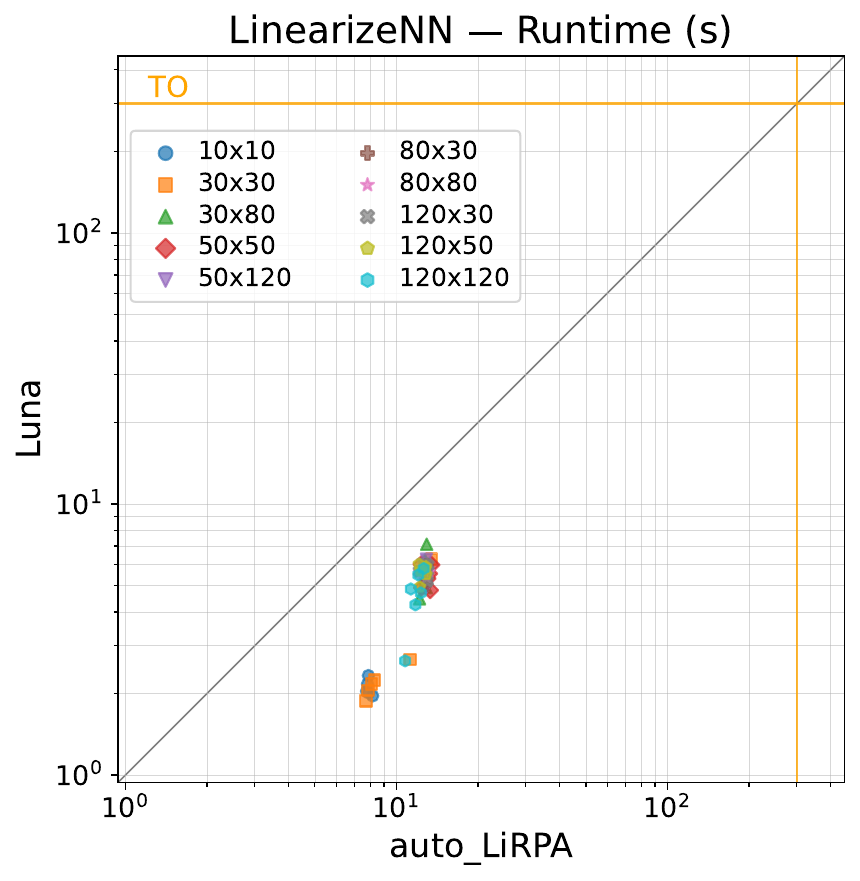}
    \caption{Bound width (left) and runtime (right) for Linearizenn 2024 (GPU)}
    \label{fig:gpu_linearizenn_2024}
\end{figure}

\begin{figure}[!ht]
    \centering
    \includegraphics[width=0.33\linewidth]{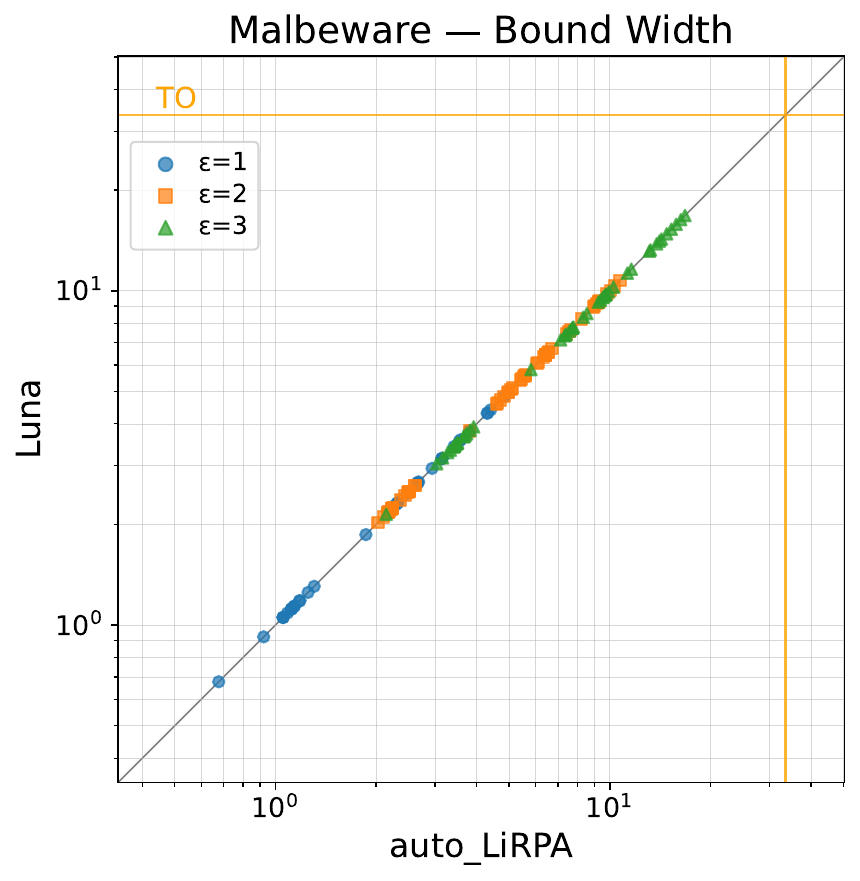}
    \hfil
    \includegraphics[width=0.33\linewidth]{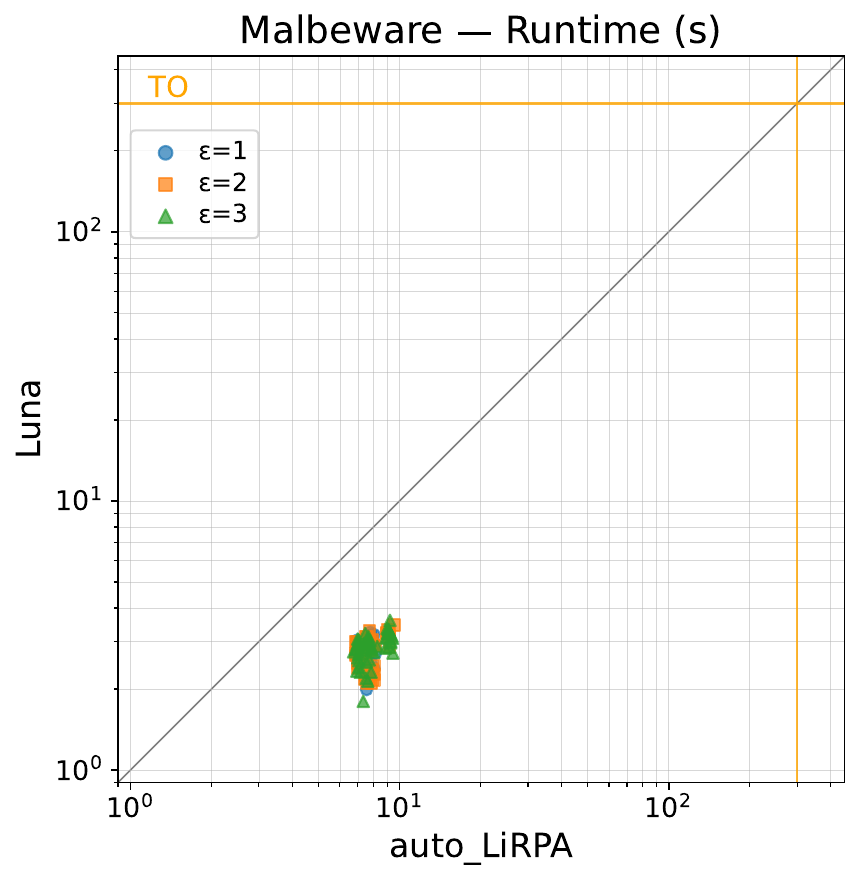}
    \caption{Bound width (left) and runtime (right) for Malbeware (GPU)}
    \label{fig:gpu_malbeware}
\end{figure}

\begin{figure}[!ht]
    \centering
    \includegraphics[width=0.33\linewidth]{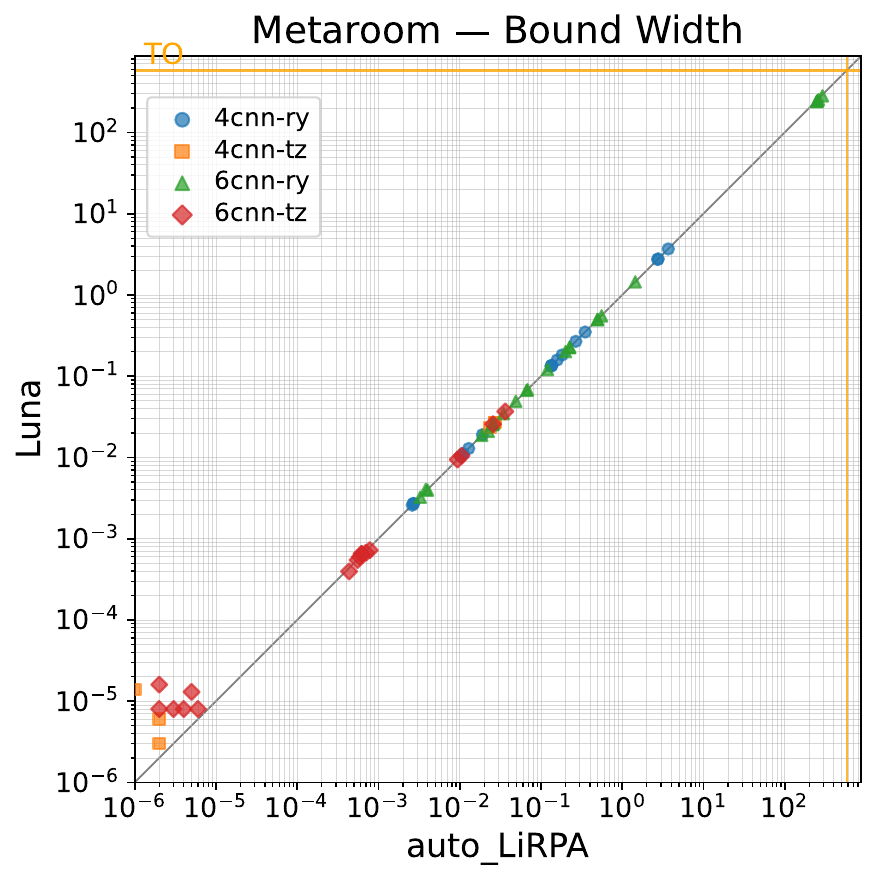}
    \hfil
    \includegraphics[width=0.33\linewidth]{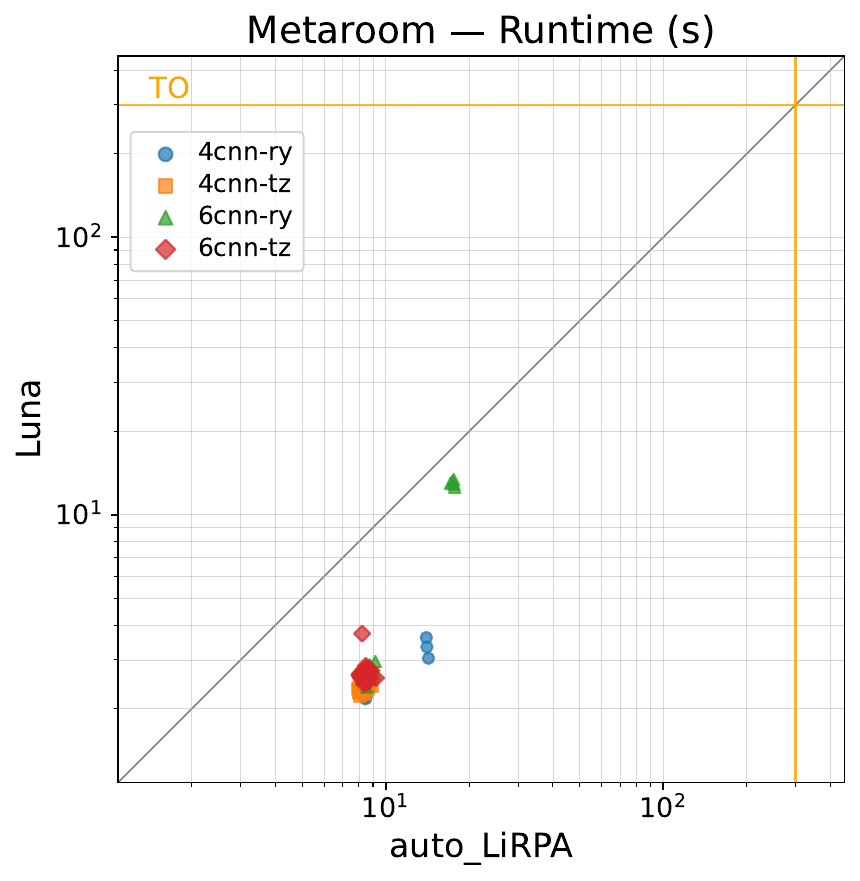}
    \caption{Bound width (left) and runtime (right) for Metaroom 2023 (GPU)}
    \label{fig:gpu_metaroom_2023}
\end{figure}

\begin{figure}[!ht]
    \centering
    \includegraphics[width=0.33\linewidth]{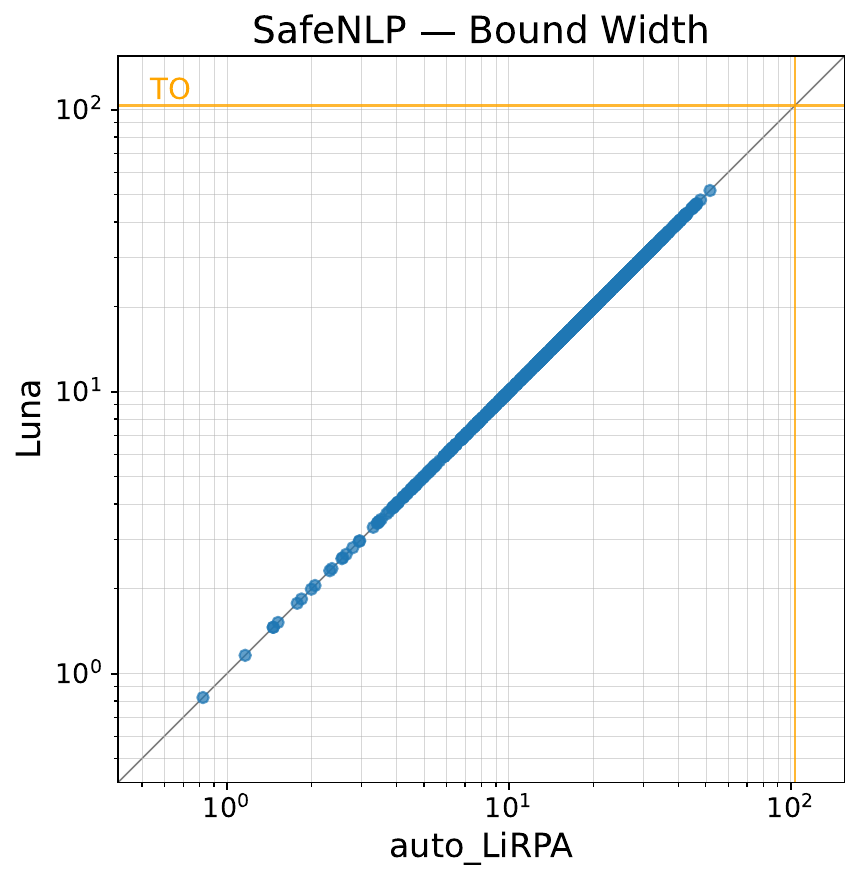}
    \hfil
    \includegraphics[width=0.33\linewidth]{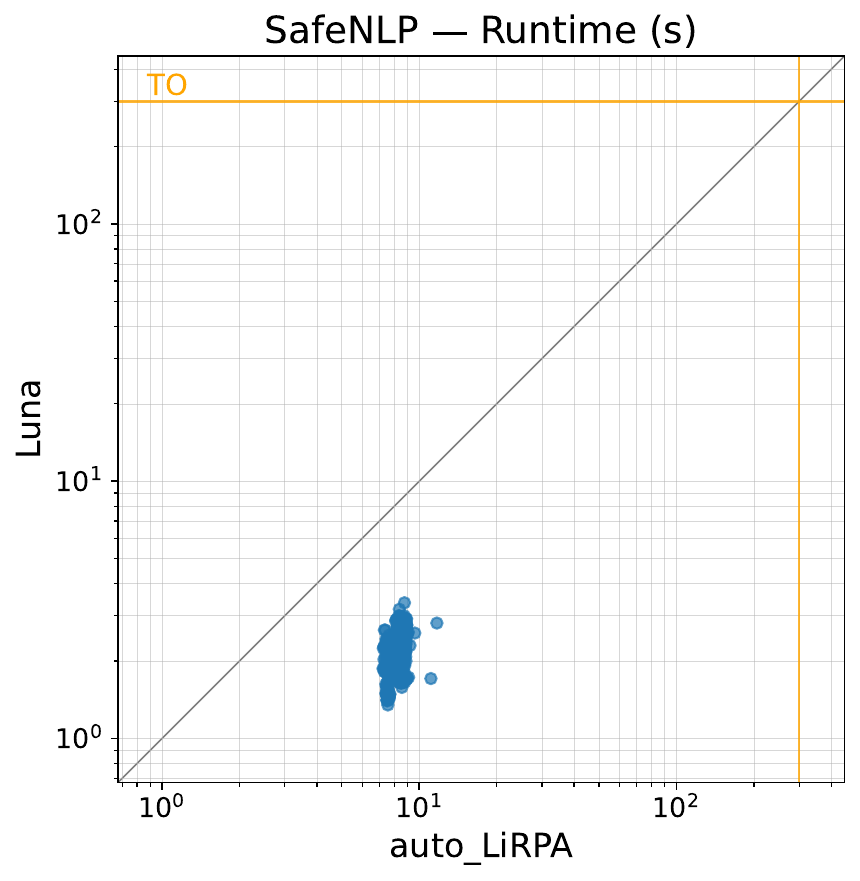}
    \caption{Bound width (left) and runtime (right) for Safenlp 2024 (GPU)}
    \label{fig:gpu_safenlp_2024}
\end{figure}

\begin{figure}[!ht]
    \centering
    \includegraphics[width=0.33\linewidth]{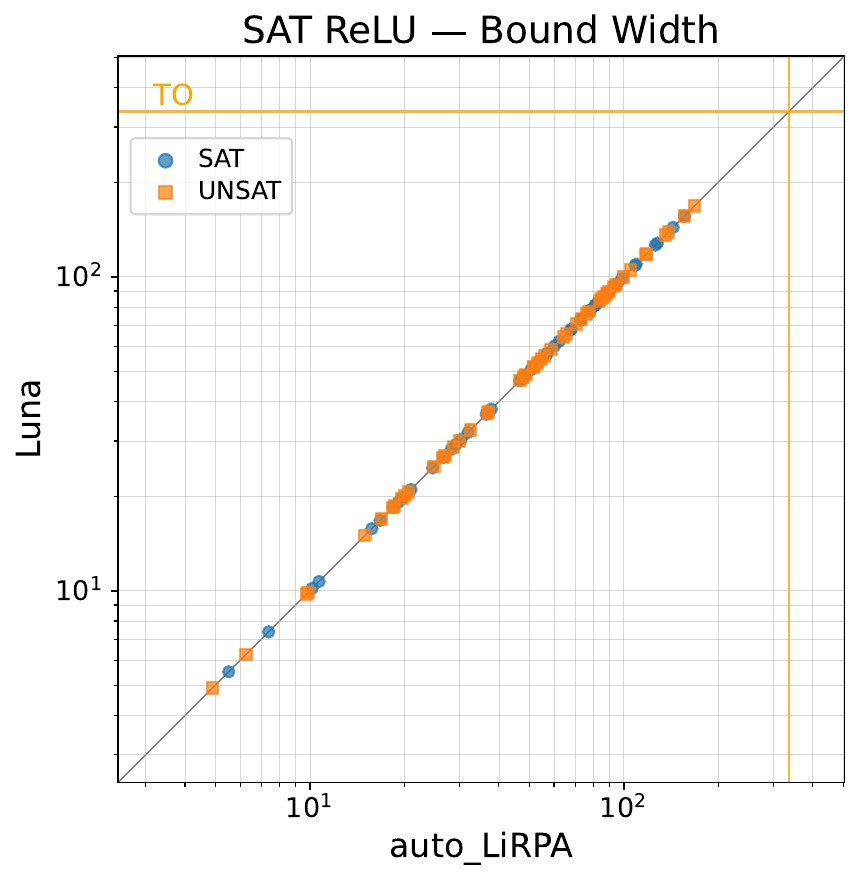}
    \hfil
    \includegraphics[width=0.33\linewidth]{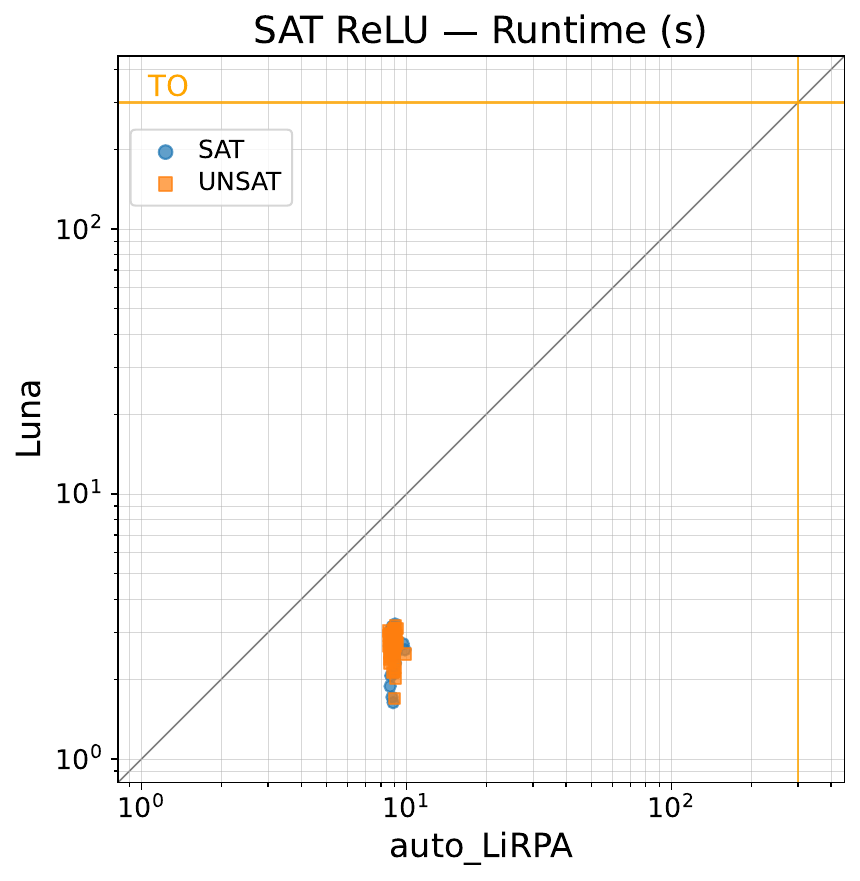}
    \caption{Bound width (left) and runtime (right) for Sat Relu (GPU)}
    \label{fig:gpu_sat_relu}
\end{figure}

\begin{figure}[!ht]
    \centering
    \includegraphics[width=0.33\linewidth]{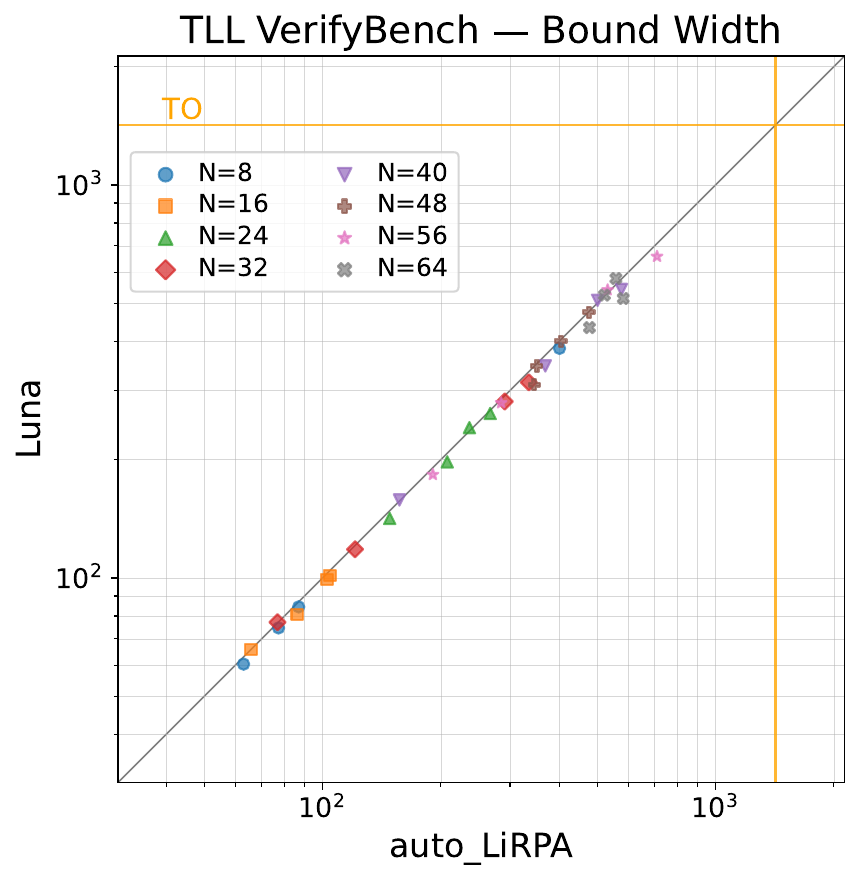}
    \hfil
    \includegraphics[width=0.33\linewidth]{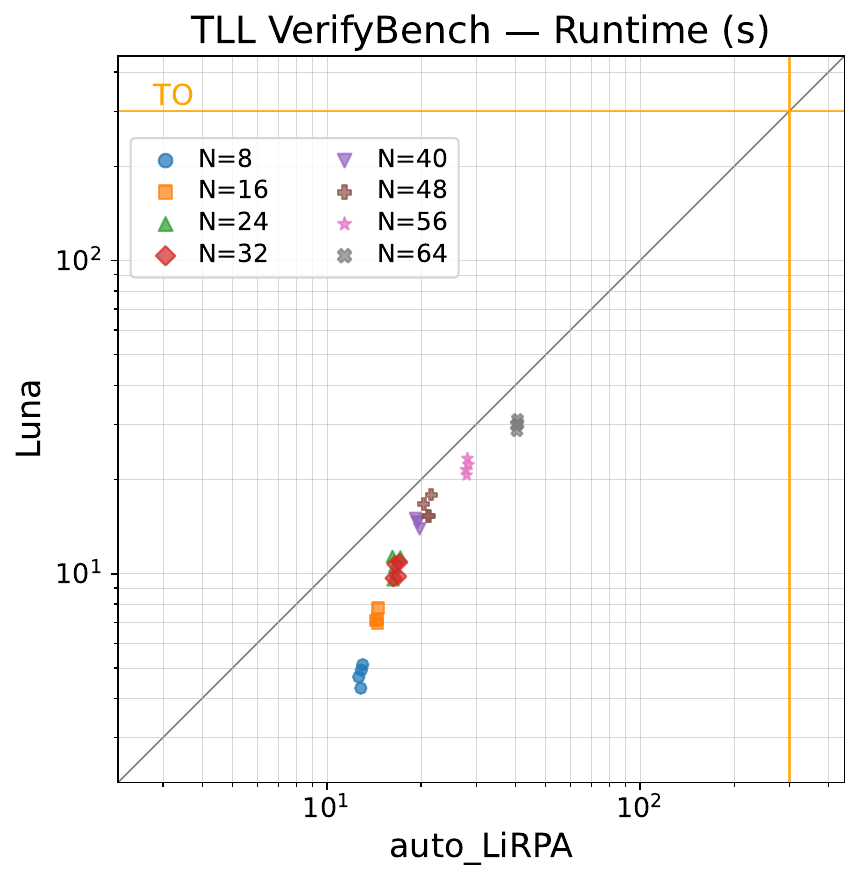}
    \caption{Bound width (left) and runtime (right) for Tllverifybench 2023 (GPU)}
    \label{fig:gpu_tllverifybench_2023}
\end{figure}

\Cref{fig:gpu_acasxu_2023,fig:gpu_cersyve,fig:gpu_collins_rul_cnn_2022,fig:gpu_cora_2024,fig:gpu_linearizenn_2024,fig:gpu_malbeware,fig:gpu_metaroom_2023,fig:gpu_safenlp_2024,fig:gpu_sat_relu,fig:gpu_tllverifybench_2023} show patterns very similar to their CPU counterparts in \Cref{fig:acasxu_2023,fig:cersyve,fig:collins_rul_cnn_2022,fig:cora_2024,fig:linearizenn_2024,fig:malbeware,fig:metaroom_2023,fig:safenlp_2024,fig:sat_relu,fig:tllverifybench_2023}. On both hardware settings the relative behavior of the two tools is the same: \sys and \autolirpa produce comparable bound widths on the same instances, and \sys is consistently faster. The main difference is that GPU acceleration eliminates the timeouts seen on CPU; both tools now complete every instance across every benchmark, and \sys is faster on every one.

\FloatBarrier
\subsubsection{Startup Time Network Values}
\label{app:startup-time-networks}

\begin{table}[h]
\centering
\setlength{\abovecaptionskip}{10pt}
\begin{tabular}{lcccc}
Network & Layers & Activation & \# Params & Input Dim \\
\midrule
Nano    & 1 & ReLU & 1 & 1 \\
Tiny    & 2 & ReLU & 4 & 1 \\
\bottomrule
\end{tabular}
\caption{Test networks used in startup time experiments.}
\label{tab:startup_networks}
\end{table}

Table~\ref{tab:startup_networks} contains per network values for the startup time evaluation.

\newpage
\bibliographystyle{splncs04}
\bibliography{references}

\end{document}